%% file: preprint-main.tex
\RequirePackage[svgnames]{xcolor}

\documentclass[11pt,letterpaper]{mystyle}
\usepackage[all]{hypcap}
\usepackage[svgnames]{xcolor}
\usepackage[comma,authoryear,compress]{natbib}
\usepackage{hyperref}[citecolor=lightblue]
\usepackage{amsmath}
\usepackage{etoolbox}

\hypersetup{
    colorlinks = true,
    citecolor = {YaleBlue},
}

\usepackage{algorithm}
\usepackage{algorithmicx}
\usepackage{algpseudocode}
\usepackage{microtype}
\usepackage{graphicx}
\expandafter\def\csname ver@subfig.sty\endcsname{}
\usepackage{booktabs} %
\usepackage{float}
\usepackage{bigstrut}

\usepackage{comment}
\usepackage{amsmath}
\usepackage{amssymb}
\usepackage{mathtools}
\usepackage{amsthm}
\usepackage{mathrsfs}
\usepackage{nicefrac}
\usepackage{dsfont}
\usepackage{enumitem}
\usepackage{subcaption}
\usepackage{graphicx,subfig}
\usepackage{cleveref}
\usepackage{bxcoloremoji}
\usepackage{float}
\usepackage{algpseudocode}
\usepackage{xspace}

\usepackage[utf8]{inputenc} %
\usepackage[T1]{fontenc}    %
\usepackage{url}            %
\usepackage{booktabs}       %
\usepackage{amsfonts}       %
\usepackage{nicefrac}       %
\usepackage{microtype}      %
\usepackage{graphicx}
\usepackage{amssymb}
\usepackage{fdsymbol}
\usepackage{wrapfig}
\usepackage{lipsum}
\usepackage{enumitem}
\usepackage{stackengine}
\usepackage[font=small,labelfont=bf]{caption}
\usepackage{color}
\usepackage{adjustbox}

\usepackage{rotating}
\usepackage{makecell}

\usepackage{multirow}
\usepackage{pdfpages}

\input{macro}

\newtcolorbox{AIbox}[2][]{aibox,title=#2,#1}
\definecolor{lightblue}{rgb}{0.22,0.45,0.70}%
\definecolor{Gray}{gray}{0.95}
\definecolor{Cornsilk}{rgb}{1.0, 0.97, 0.86}

\usepackage{url}
\usepackage{pifont}
\usepackage{soul}

\usepackage{amsmath}
\usepackage[all]{hypcap}

\newcommand{\taxonomy}{\textsc{GameUI-Taxonomy}\xspace}
\newcommand{\engine}{\textsc{G2WEngine}\xspace}
\newcommand{\GameDataset}{\textsc{Game2World}\xspace}
\newcommand{\GameDatasetS}{\textsc{Game2World-S}\xspace}
\newcommand{\GameDatasetW}{\textsc{Game2World-W}\xspace}
\newcommand{\method}{\textsc{GameCleaner}\xspace}

\usepackage{algpseudocode}
\usepackage{authblk}   
\usepackage{amssymb}      
\usepackage{bbm}          
\usepackage{optidef}  
\usepackage[utf8]{inputenc}
\usepackage[T1]{fontenc}

\usepackage{mathpazo}
\usepackage{tcolorbox}
\tcbuselibrary{skins,breakable,listings} 

\usepackage{lipsum}

\input{preamble}

\newtcolorbox{simpleElegantQuote}{
    colback=AliceBlue!50!White,   
    colframe=RoyalBlue!75!Black,  
    boxrule=0.5pt,                
    arc=2mm,                     
    boxsep=4pt,                   
    left=10pt, right=10pt,        
    top=8pt, bottom=8pt,         
    fontupper=\itshape,          
}

\title{\textsc{Game2World} Engine: Unlocking In-the-Wild Gameplay Videos for World Model Training}

\runningtitle{}

\author{%
 Wenxuan Shen, Dongna Jin, Dongping Chen\textsuperscript{$\ddagger$} \\

}

\correspondingauthor{Dongping Chen: dongpingchen0612@gmail.com.}

\begin{document}

\begin{figure*}[!b]
    \centering
    \vspace{-2em}
    \includegraphics[width=1\linewidth]{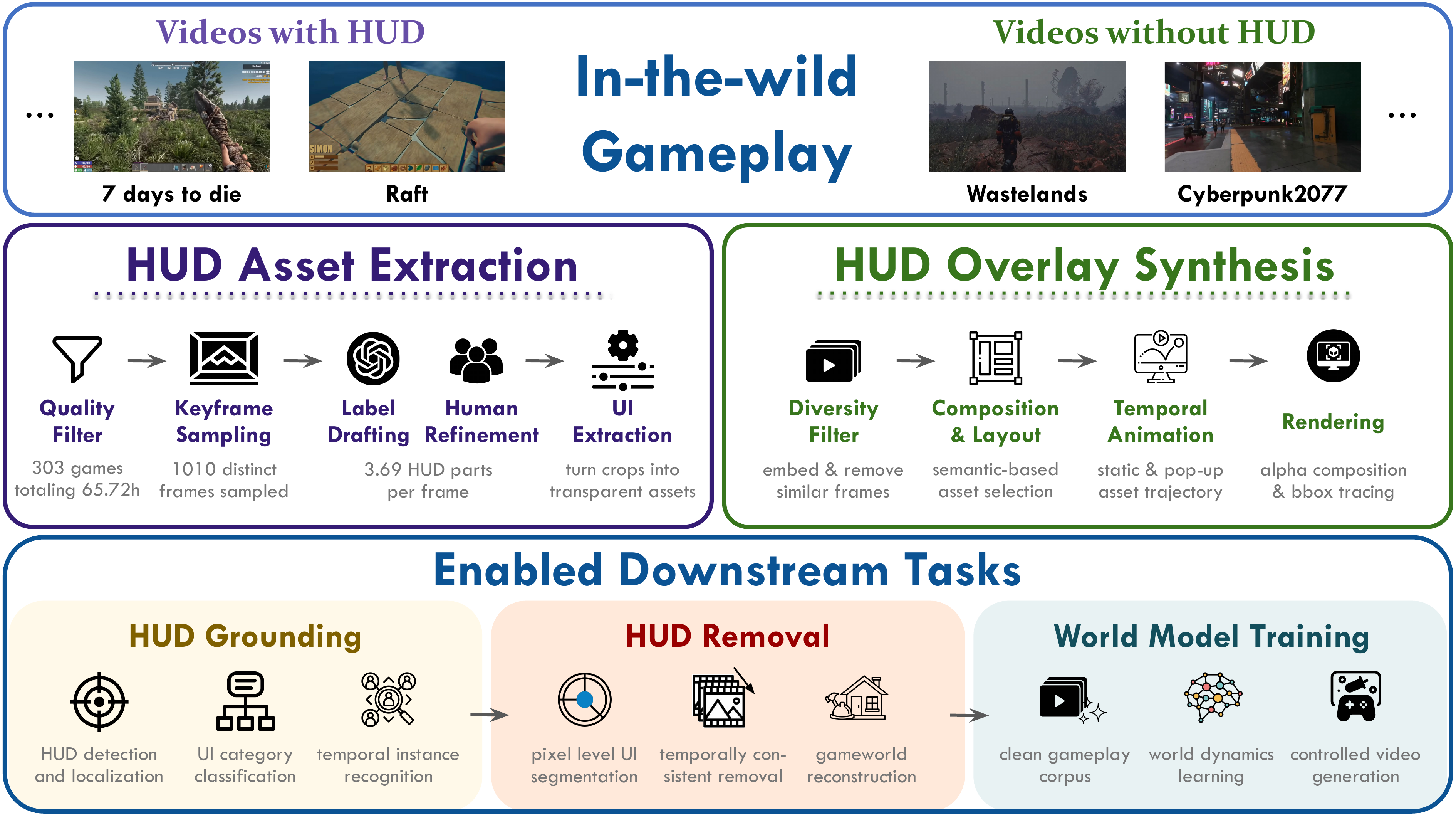}
    \caption{Overview of \engine, a scalable data engine that transforms in-the-wild gameplay videos into world-model-ready data through HUD asset extraction, temporally coherent UI synthesis, and downstream support for HUD understanding, removal, and clean gameplay generation.}
    \label{fig:dataset_pipeline}
\end{figure*}

\input{Sec/0-abstract}

\maketitle

\input{Sec/1-intro}

\input{Sec/2-related}
\input{Sec/3-pilot}

\input{Sec/4-dataset}
\input{Sec/5-benchmark}
\input{Sec/6-method}
\input{Sec/7-experiment}
\input{Sec/8-discussion}
\input{Sec/10-conclusion}

%%%%%%%%%%%%%%%%%%%%%%%%%%%%%%%%%%%%%%%%%%%%%%%%%%%%%%%%%%%%

{
    \small
    \bibliography{aaai2026}
}

\clearpage
\appendix
% \input{checklist}
\input{Sec/99-appendix}

%%%%%%%%%%%%%%%%%%%%%%%%%%%%%%%%%%%%%%%%%%%%%%%%%%%%%%%%%%%%

% \input{checklist.tex}

\end{document}

%% file: macro.tex
\definecolor{blanchedalmond}{rgb}{1.0, 0.92, 0.8}
\definecolor{carmine}{rgb}{0.59, 0.0, 0.09}
\definecolor{lightblue}{rgb}{0.22,0.45,0.70}%

\renewcommand{\mathbf}{\boldsymbol}

\makeatletter
\def\Ddots{\mathinner{\mkern1mu\raise\p@
\vbox{\kern7\p@\hbox{.}}\mkern2mu
\raise4\p@\hbox{.}\mkern2mu\raise7\p@\hbox{.}\mkern1mu}}
\makeatother

\definecolor{amaranth}{rgb}{0.9, 0.17, 0.31}
\definecolor{antiquebrass}{rgb}{0.8, 0.58, 0.46}
\definecolor{antiquefuchsia}{rgb}{0.57, 0.36, 0.51}
\definecolor{chromeyellow}{rgb}{0.31, 0.47, 0.26}

%% file: preamble.tex
\usepackage[utf8]{inputenc}
\usepackage[T1]{fontenc}
\usepackage{microtype}

\usepackage{amsmath}
\usepackage{amsfonts}
\usepackage{amssymb}
\usepackage{nicefrac}

\usepackage[dvipsnames,svgnames,x11names,table]{xcolor}

\usepackage{graphicx}
\usepackage{caption}
\usepackage{subcaption}
\usepackage{wrapfig}
\usepackage{placeins}
\usepackage{cuted}
\usepackage{balance}

\usepackage{booktabs}
\usepackage{multirow}
\usepackage{makecell}
\usepackage{tabularx}
\usepackage{array}
\usepackage{adjustbox}
\usepackage{siunitx}
\usepackage{wrapfig}
\usepackage{graphicx}
\usepackage{enumerate}
\usepackage{enumitem}

\usepackage{fancyvrb}
\usepackage{listings}
\usepackage{mdframed}
\usepackage{tcolorbox}

\usepackage{xspace}
\usepackage{soul}
\usepackage{ragged2e}
\usepackage{caption}
\usepackage{float}
\usepackage{pifont}
\usepackage{wasysym}

\definecolor{deeppink}{RGB}{231,84,128}

\usepackage{hyperref}

\newcounter{finding}

\newtcbox{\toolcalltag}{
  on line,
  arc=0pt,
  boxrule=0.4pt,
  colback=blue!8,
  colframe=blue!60!black,
  left=2pt,
  right=2pt,
  top=1pt,
  bottom=1pt,
  boxsep=1pt,
  fontupper=\small
}

\newtcbox{\toolresptag}{
  on line,
  arc=0pt,
  boxrule=0.4pt,
  colback=green!8,
  colframe=green!50!black,
  left=2pt,
  right=2pt,
  top=1pt,
  bottom=1pt,
  boxsep=1pt,
  fontupper=\small
}

\newtcbox{\finalanswertag}{
  on line,
  arc=0pt,
  boxrule=0.4pt,
  colback=red!8,
  colframe=red!60!black,
  left=2pt,
  right=2pt,
  top=1pt,
  bottom=1pt,
  boxsep=1pt,
  fontupper=\small
}

\newtcbox{\divergetag}{
  on line,
  arc=0pt,
  boxrule=0.4pt,
  colback=orange!10,
  colframe=orange!70!black,
  left=2pt,
  right=2pt,
  top=1pt,
  bottom=1pt,
  boxsep=1pt,
  fontupper=\small
}

\newtcolorbox{sharedprefixbox}{
  enhanced,
  breakable,
  colback=gray!4,
  colframe=black!35,
  boxrule=0.5pt,
  arc=0pt,
  left=4pt,
  right=4pt,
  top=4pt,
  bottom=4pt,
  fontupper=\small,
  fonttitle=\small\bfseries
}

\newtcolorbox{failedtrajbox}{
  enhanced,
  breakable,
  colback=red!2,
  colframe=red!55!black,
  boxrule=0.6pt,
  arc=0pt,
  left=4pt,
  right=4pt,
  top=4pt,
  bottom=4pt,
  title=\textbf{Failed Trajectory},
  colbacktitle=red!10,
  coltitle=black,
  fontupper=\small,
  fonttitle=\small\bfseries
}

\newtcolorbox{successtrajbox}{
  enhanced,
  breakable,
  colback=green!2,
  colframe=green!45!black,
  boxrule=0.6pt,
  arc=0pt,
  left=4pt,
  right=4pt,
  top=4pt,
  bottom=4pt,
  title=\textbf{Successful Trajectory},
  colbacktitle=green!10,
  coltitle=black,
  fontupper=\small,
  fonttitle=\small\bfseries
}

\newtcblisting{agentpromptbox}[2][]{%
  enhanced,
  listing only,
  listing options={
    basicstyle=\ttfamily\scriptsize,
    columns=fullflexible,
    breaklines=true,
    breakatwhitespace=false,
    keepspaces=true,
    showstringspaces=false,
    upquote=true
  },
  attach boxed title to top center={yshift=-3mm,yshifttext=-1mm},
  boxrule=0.9pt,
  colback=gray!00,
  colframe=black!50,
  colbacktitle=gray!25,
  boxed title style={size=small,colframe=gray!50},
  title={#2},
  #1
}

%% file: Sec/0-abstract.tex
\begin{abstract}
Video games provide a scalable source of training data for video world models, offering diverse environments, complex interactions, and abundant in-the-wild gameplay videos. However, raw gameplay footage entangles the game world with screen-space interfaces, introducing game-specific biases and irrelevant dynamics that hinder world-model training. To address this problem, we introduce \taxonomy and \engine, a full-stack framework that formalizes gameplay UI grounding and removal. \engine automatically extracts reusable UI assets from real gameplay videos and synthesizes temporally coherent UI overlays on clean footage. Using this engine, we construct \GameDataset, comprising 96K synthetic paired videos with precise reconstruction targets and 1,079 in-the-wild clips from 303 games for realistic evaluation. Its asset library contains 5,132 verified UI elements across 21 taxonomy categories, collected from 1,010 representative gameplay frames. Based on \GameDataset, we propose \method, a mask-free gameplay UI removal model that combines multimodal semantic understanding with video editing capabilities. Unlike mask-based methods, \method directly identifies and removes diverse HUD elements while preserving the underlying scene content and temporal dynamics. In a controlled pilot, world models trained on UI-free gameplay improve overall VideoReward by 6.83\% over those trained on UI-overlaid data. On UI-removal evaluation, \method achieves an average AAR of 95.36 on synthetic videos, outperforming the strongest temporal mask baseline by 57.3\%, and obtains the best in-the-wild AAR of 80.05 with 99.8 background preservation. These results demonstrate the scalable potential of transforming Internet gameplay videos into high-quality world-model training data. Code, dataset and model will be available at \href{https://github.com/Dongping-Chen/Game2World}{\texttt{Dongping-Chen/Game2World}}.

\end{abstract}

%% file: Sec/1-intro.tex
\section{Introduction}

World models have emerged as a promising foundation for interactive simulation, controllable video generation, and embodied agents. However, scaling them remains challenging because collecting diverse, long-horizon interaction data from closed simulators or instrumented environments is expensive and difficult to generalize across domains. In-the-wild gameplay videos offer a scalable alternative: the Internet contains abundant footage spanning diverse 3D environments, visual styles, viewpoints, tasks, dynamics, and player behaviors, providing rich visual experience for learning how complex environments evolve over time.

However, raw gameplay footage is not a direct observation of the game world, but a composite rendering that mixes the underlying scene with screen-space HUDs, menus, notifications, watermarks, and streaming overlays \citep{kang2020instance,kang2022unsupervised,wu2023widget}. These elements introduce game-specific patterns and dynamics unrelated to the physical environment. In our controlled pilot study, we train otherwise identical video-generation models on clean and UI-overlaid versions of the same 5,442 gameplay clips. Removing UI improves overall VideoReward by 6.83\%, including gains of 18.8\% in motion quality and 2.7\% in video quality. Clean clips also achieve 8.59\% higher aesthetic quality and 4.36\% stronger motion consistency, showing that gameplay interfaces introduce irrelevant supervision and visual shortcuts that reduce the value of gameplay videos for world-model training.

To unlock this data source, we first formalize gameplay UI understanding through a unified \taxonomy covering diverse interface elements and supporting tasks such as UI grounding and removal. We then introduce \engine, a processing framework that transforms raw gameplay footage into world-model-ready data. We formulate HUD removal as \textbf{interface--world disentanglement}: identifying structured screen-space overlays, removing them consistently over time, and reconstructing occluded game content while preserving unrelated scene information. Unlike physical objects, gameplay interfaces exhibit category-specific spatial and temporal behaviors, requiring gameplay-aware semantic understanding and temporally coherent reconstruction.

Based on \engine, we construct \GameDataset, which combines 96K synthetic UI-overlaid videos with clean references and 1,079 in-the-wild clips containing authentic gameplay interfaces. The two subsets evaluate complementary aspects of the task: faithful reconstruction under controlled ground truth and generalization to real gameplay footage. Building on \GameDataset, we propose \method, a mask-free gameplay UI removal model combining multimodal semantic understanding with video editing. \method directly identifies and removes diverse HUD elements while reconstructing occluded content and preserving unrelated scene details and temporal dynamics, enabling generalization across unseen games and UI layouts.

Across controlled and in-the-wild evaluations, \method achieves a stronger balance between UI removal and game-world preservation than existing video editing and object removal methods. On the synthetic subset, it obtains an average score of 0.5697, outperforming the strongest mask-free baseline, LoomVideo, by 40.3\%, while remaining only 0.0233 below mask-assisted EffectErase. On the more challenging in-the-wild subset, \method achieves the highest overall score of 0.4886, exceeding EffectErase by 23.7\%. Together with our pilot study, these results show that gameplay UI removal improves the utility of in-the-wild gameplay footage as training data rather than serving only as a visual editing operation.

Our main contributions are:
\begin{itemize}
    \item \textbf{Pilot study of gameplay video as world-model training data.}
    We systematically isolate the effect of gameplay interfaces using clean and UI-overlaid versions of the same 5,442 clips. Training on clean gameplay improves overall VideoReward by 6.83\%, motion quality by 18.8\%, aesthetic quality by 8.59\%, and motion by 4.36\%.

    \item \textbf{Full-stack taxonomy, task formulation, and evaluation protocol.}
    We introduce a unified taxonomy of gameplay HUD elements, formalize HUD grounding and removal, and develop element-level removal, artifact, and background-preservation metrics. Our MLLM evaluator reaches an 87.78\% removed F1 and a Cohen's $\kappa$ of 0.820 against human annotations.

    \item \textbf{Scalable data engine and large-scale dataset.}
    We develop \textsc{G2WEngine} for extracting real HUD assets and synthesizing temporally coherent paired videos. It produces \textsc{Game2World}, containing 96K synthetic video pairs, 1,079 in-the-wild clips from 303 games, and 5,132 verified assets across 21 UI categories.

    \item \textbf{\textsc{GameCleaner}: mask-free gameplay UI removal.}
    We develop a general gameplay video editing model that removes diverse overlays without input masks. It achieves 95.36 on the synthetic dataset, outperforming the strongest mask-free baseline by 57.3\%, and obtains the best in-the-wild AAR score of 80.05 (w. ref) and 51.60 (w.o. ref).
\end{itemize}

%% file: Sec/2-related.tex
\section{Related Work}
\subsection{Learning World Models from Gameplay Videos}
Video games provide controllable environments and abundant trajectories for learning world dynamics. Early work learned compact action-conditioned simulators from Atari interactions \citep{micheli2022transformers}, while DIAMOND used diffusion models to construct reinforcement-learning environments and interactive neural game engines from recorded gameplay \citep{alonso2024diffusion}. Genie extended this direction to unlabeled Internet videos by jointly learning latent actions and environment dynamics \citep{bruce2024genie, genie3}. Recent systems further train action-conditioned generative models to simulate interactive 3D worlds \citep{valevski2025diffusion,che2025gamegen,guo2025mineworld,zhang2025matrix}. However, they generally treat rendered frames as direct world observations without separating game content from overlays such as HUDs and menus.

\subsection{Video Inpainting for Object Removal}
Video inpainting removes unwanted objects while reconstructing spatially plausible and temporally consistent content. Early methods used 3D convolutions or optical-flow completion \citep{chang2019free,xu2019deep}, while later approaches introduced spatial-temporal attention, long-range feature matching, and motion-guided propagation, including STTN, E2FGVI, and ProPainter \citep{zeng2020learning,li2022towards,zhou2023propainter}. Recent work explores language-conditioned removal, blind inpainting, and diffusion models for large missing regions and associated effects such as shadows or reflections \citep{wu2024towards,wu2025bvinet,lee2024generative}. Nevertheless, these methods mainly target physical objects in natural videos, whereas gameplay UIs are structured screen-space overlays with distinct spatial and temporal patterns.

%% file: Sec/3-pilot.tex
\section{Pilot Study: UI-Removed Gameplay Videos Are Great Training Data for World Model}

\begin{table*}[t]
\centering

\begin{minipage}[t]{0.43\textwidth}
\vspace{0pt}
\centering
\captionof{table}{VideoReward scores of the original training data and generated videos. Higher scores are better.}
\label{tab:clean_UI-overlayed_videoreward}
\resizebox{\linewidth}{!}{
\begin{tabular}{lcccc}
\toprule[1.5pt]
\textbf{Data Type}
& \textbf{VQ}
& \textbf{MQ}
& \textbf{TA}
& \textbf{Overall} \\
\midrule
\multicolumn{5}{l}{\textit{Training Data}} \\ 
Clean
& -0.877
& \textbf{-0.644}
& \textbf{0.499}
& \textbf{-1.022} \\
UI-overlayed
& \textbf{-0.807}
& -0.704
& 0.450
& -1.061 \\
\midrule
\multicolumn{5}{l}{\textit{Finetuned Wan-2.1-T2V-1.3B Model}} \\
Clean
& \textbf{-0.643}
& \textbf{-0.298}
& 1.008
& \textbf{0.068} \\
UI-overlayed
& -0.661
& -0.367
& \textbf{1.018}
& -0.009 \\
$\Delta$
& \textcolor{red}{+2.7\%}
& \textcolor{red}{+18.8\%}
& \textcolor{green}{-1.0\%}
& \textcolor{red}{\textbf{+6.83\%}} \\
\bottomrule[1.5pt]
\end{tabular}
}
\end{minipage}
\hfill
\begin{minipage}[t]{0.55\textwidth}
\vspace{0pt}
\centering
\captionof{table}{Comparison of visual quality and motion statistics between clean and UI-overlayed videos.}
\label{tab:clean_UI-overlayed_quality}
\resizebox{\linewidth}{!}{
\begin{tabular}{lccc}
\toprule[1.5pt]
\textbf{Data Type}
& \textbf{Clarity (MUSIQ)}
& \textbf{Aes. (LAION)}
& \textbf{Motion (RAFT)} \\
\midrule
Clean
& 69.254
& \textbf{6.109}
& \textbf{16.392} \\
UI-overlayed
& \textbf{70.260}
& 5.626
& 15.706 \\
$\Delta$
& \textcolor{green}{-1.43\%}
& \textcolor{red}{+8.59\%}
& \textcolor{red}{+4.36\%} \\
\bottomrule[1.5pt]
\end{tabular}
}
\end{minipage}

\end{table*}

We conduct a pilot study to examine whether high-quality in-the-wild gameplay videos can support video-based world-model training. Although gameplay footage provides diverse environments, dynamics, and long-horizon trajectories, it often contains UIs, menus, notifications, watermarks, and streaming overlays unrelated to the game world. We investigate whether these non-world elements reduce the quality of training data for generative world models. We use a text-to-video model as a proxy, following recent gameplay world models built on video-generation backbones such as Wan. Since our focus is data quality, we do not train action-conditioning modules.

\paragraph{Experiment Setups.}
To isolate the impact of UI, we collect 5,442 UI-free gameplay clips and construct two training sets from identical videos: original clean clips and UI-overlaid clips synthesized by \textsc{Game2WorldEngine}. We generate captions using Qwen3.5-9B \citep{qwen3.5} and fine-tune \texttt{Wan2.1-T2V-1.3B} on each dataset. We evaluate both training data and generated videos using VideoReward \citep{liu2025improving}, VBench prompts \citep{huang2024vbench}, and direct video quality metrics including clarity, aesthetics, and motion \citep{wang2025koala36m}.

\paragraph{Empirical Results.}
As shown in Tables~\ref{tab:clean_UI-overlayed_videoreward} and~\ref{tab:clean_UI-overlayed_quality}, clean gameplay videos consistently provide better training signals than UI-overlaid counterparts. After fine-tuning, models trained on clean videos improve the overall VideoReward by \textbf{6.83\%}, with larger gains in motion quality (+18.8\% MQ) and video quality (+2.7\% VQ), while achieving comparable temporal alignment. Clean videos also improve aesthetic quality by \textbf{8.59\%} and motion consistency by \textbf{4.36\%}, despite a small decrease in low-level clarity (-1.43\%), likely because UI overlays contain sharp text and interface graphics favored by clarity metrics. Our clean gameplay data reaches a LAION aesthetic score of 6.109, surpassing Panda-70M and Koala-36M~\citep{wang2025koala36m}. These results demonstrate that removing UI improves the effectiveness of gameplay videos as training data for world models.

%% file: Sec/4-dataset.tex
\section{\engine: Constructing World Model Data from In-the-Wild Gameplay Videos}

As illustrated in Fig \ref{fig:dataset_pipeline}, we introduce \engine, an automatic data engine that transforms raw gameplay videos into structured training data for gameplay UI understanding and removal. The core challenge is that gameplay videos entangle the underlying game world with screen-space interfaces, whose spatial layouts, semantic meanings, and temporal behaviors vary substantially across games. \engine therefore performs interface--world disentanglement through four stages: defining a unified UI taxonomy, extracting reusable UI assets from real gameplay videos, curating clean gameplay videos, and synthesizing temporally coherent UI overlays with complete supervision.

\begin{figure*}[!t]
    \centering
    \includegraphics[width=1\linewidth]{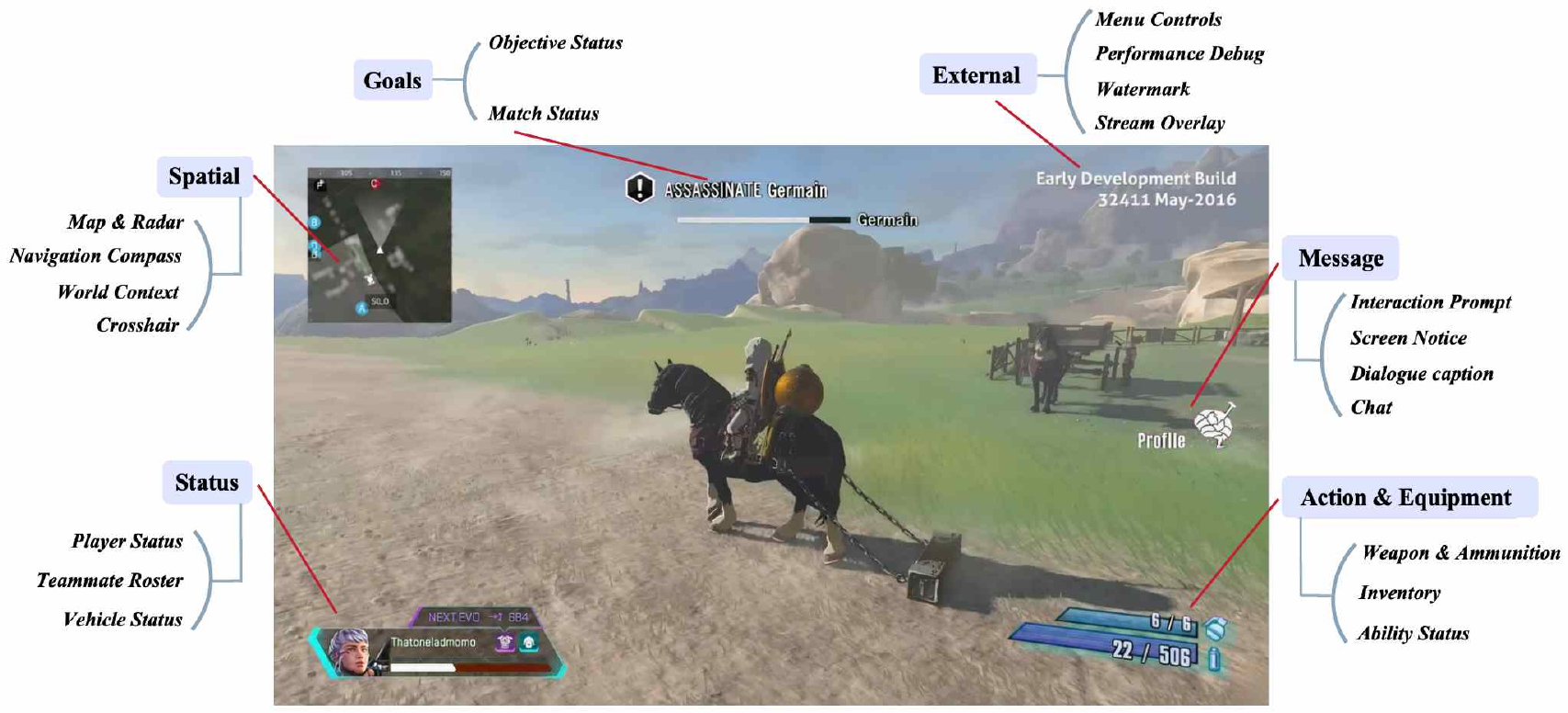}
    \caption{Overview of \taxonomy, which organizes gameplay HUD elements into categories based on their functional roles and rendering behaviors, including spatial information, player status, goals, messages, actions, and external overlays.}
    \label{fig:taxonomy}
\end{figure*}

\begin{figure*}
    \centering
    \includegraphics[width=0.95\linewidth]{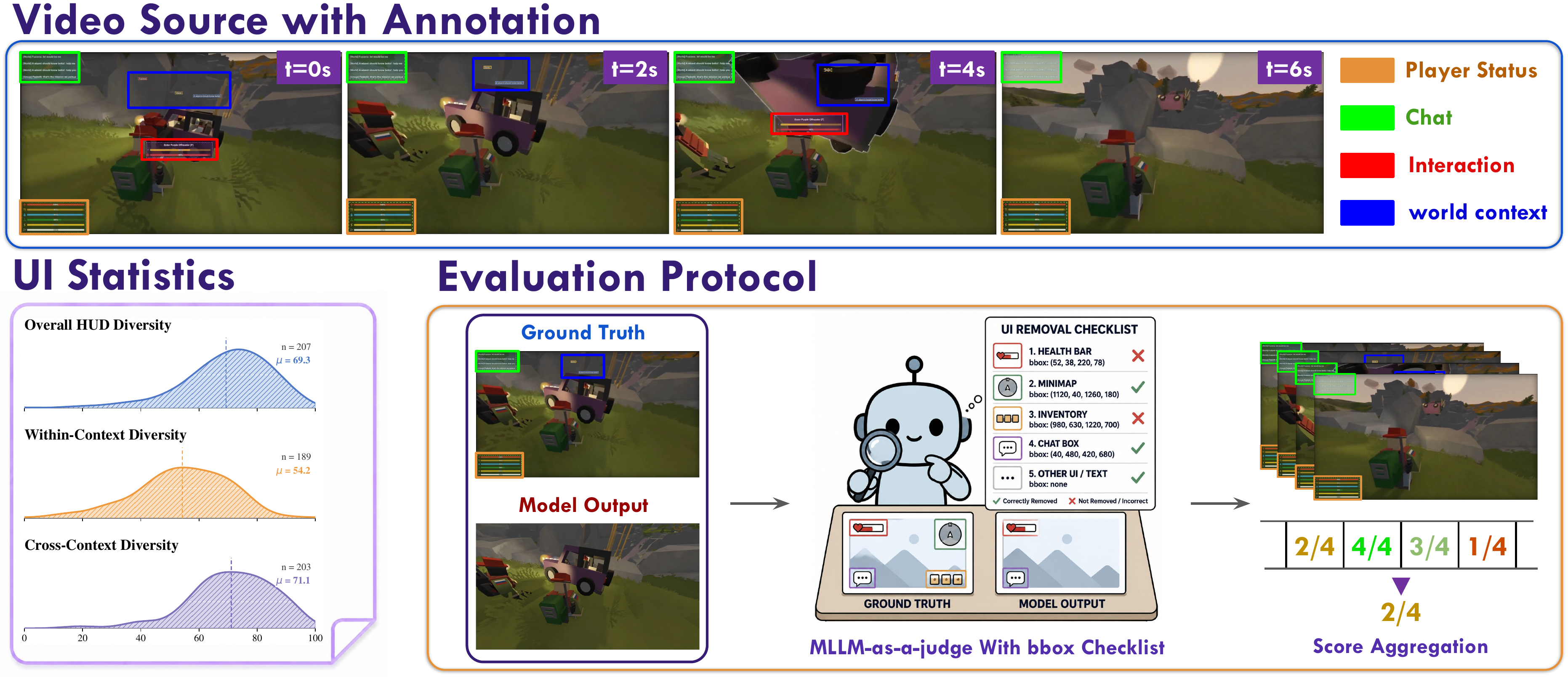}
    \caption{Overview of the Game2World-W dataset. The dataset provides temporally annotated in-the-wild gameplay videos, and evaluates UI removal through an MLLM-as-a-judge protocol with element-level bounding-box checklists.}
    \label{fig:game2world-w}
\end{figure*}

\subsection{GameUI-Taxonomy}

We first define \taxonomy, a unified taxonomy that organizes gameplay UI according to their semantic roles and rendering behaviors. Unlike conventional object categories, UI elements are screen-space entities with strong priors in location, persistence, and temporal dynamics. The taxonomy enables consistent annotation, category-aware asset sampling, and controllable UI synthesis. The complete taxonomy and statistics are provided in Fig.\ref{fig:taxonomy} and Fig \ref{fig:taxonomy_stats}.

\subsection{UI Asset Extraction}

To construct a reusable UI asset library, \engine collects diverse gameplay videos from the Internet and samples representative frames containing different gameplay states, interface configurations, and visual styles. For each sampled frame, an annotation model, GPT-5.6 Terra, proposes UI bounding boxes and taxonomy labels, followed by human verification to ensure annotation quality.

Accepted UI regions are extracted from their original frames and converted into transparent assets while removing surrounding gameplay content. Each asset retains metadata including its category, source video, original location, and rendering information. These spatial and semantic annotations preserve the characteristics of real gameplay interfaces and enable realistic reconstruction during synthesis.

\subsection{Clean Gameplay Corpus Curation}

Because raw gameplay videos often contain unwanted interfaces, menus, and recording artifacts, \engine separately constructs a clean gameplay corpus for controlled UI synthesis. We filter gameplay-style videos spanning diverse genres and visual styles, segment them into 5s clips, and reduce redundancy through clustering with CLIP \citep{radford2021learning} embeddings and removing temporally adjacent clips. The resulting clips provide diverse underlying world content without interface contamination.

\subsection{UI Overlay Synthesis}

We generate paired training examples by compositing the
extracted UI assets onto clean 5-second gameplay clips. UI categories are sampled using category-specific probabilities and maximum instance counts. Transient pop-ups use a reduced long-tailed distribution to mimic realistic settings where simple and complex UI overlays both exist.

For each selected asset, the pipeline samples its rendering style, temporal duration, animation type, and motion parameters. Recorded coordinates are directly used when compatible, with a set of category-specific spatial anchor fallbacks. Persistent elements receive small coordinate perturbations for data augmentation, followed by collision detection and moderate repositioning to avoid major overlaps.

The pipeline also performs category-aware preprocessing
for structured interface components. Inventory panels, for
example, can be decomposed into reusable background bars
or grids and separately rendered item icons. This allows the
renderer to synthesize new inventory configurations while
preserving coherent layout and visual structure. The resulting generated sample contains aligned clean video, UI-overlaid video, frame-level UI masks, bounding boxes, taxonomy labels, and rendering metadata.

%% file: Sec/5-benchmark.tex
\section{\GameDataset: The Dataset}

Based on \engine, we construct \GameDataset, a dataset for gameplay UI that evaluates temporal spatial grounding and removal. 

\paragraph{Game2World-S: Synthetic UI Gameplay Dataset.} \GameDatasetS is generated by applying the UI overlay synthesis pipeline of \engine to clean gameplay videos. Its asset library is built from 1,010 representative keyframes collected from 303 games, yielding 5,132 verified UI assets across 21 taxonomy categories, including player status, maps and radars, action information, communication panels, and external overlays.

To reproduce realistic interfaces, \GameDatasetS combines static HUD components with transient elements such as notifications and interaction prompts. The pipeline generates 96k paired clean and UI-corrupted videos with frame-aligned masks, instance bounding boxes, category labels, and rendering metadata. All clips are rendered at 720p 30fps, providing exact supervision for UI grounding, segmentation, temporal localization, and removal.

\paragraph{Game2World-W: In-the-Wild Gameplay Dataset.} Real gameplay interfaces contain variations that are difficult to synthesize, including unknown layouts, transparency, dynamic transitions, and recording artifacts. We therefore construct \GameDatasetW from authentic gameplay videos across diverse games.

\GameDatasetW contains 1,079 five-second clips spanning 303 games. As the underlying clean game states are unavailable, these videos have no reference targets. We annotate the first frame using GPT-5.6 Sol \citep{openai2026gpt56}, then manually update labels and bounding boxes for dynamic UI elements in each frame to improve reliability of MLLM-as-a-judge evaluation \citep{chen2024mllm}, as shown in Fig. \ref{fig:game2world-w}.

\paragraph{Analysis of UI on Gameplay Videos.} We characterize gameplay UI using three game-level metrics: \textbf{UI-D} measures variation in UI composition, \textbf{UI-R@16} counts distinct functional UI slots within 16 frames, and \textbf{Context-UI-Lift} measures the context dependence of UI layouts. Detailed definitions are provided in the appendix. Across valid games, UI-D has a mean of 69.3, while UI-R@16 has a median of 20 and ranges from 2 to 40. Context-UI-Lift is positive for 90.8\% of games, showing that UI compositions vary more across gameplay contexts than within the same context. These results indicate that gameplay UIs are diverse, functionally rich, and strongly context-dependent.

%% file: Sec/6-method.tex
\section{\method: UI Removal Model}

\begin{figure}[!t]
    \centering
    \includegraphics[width=\linewidth]{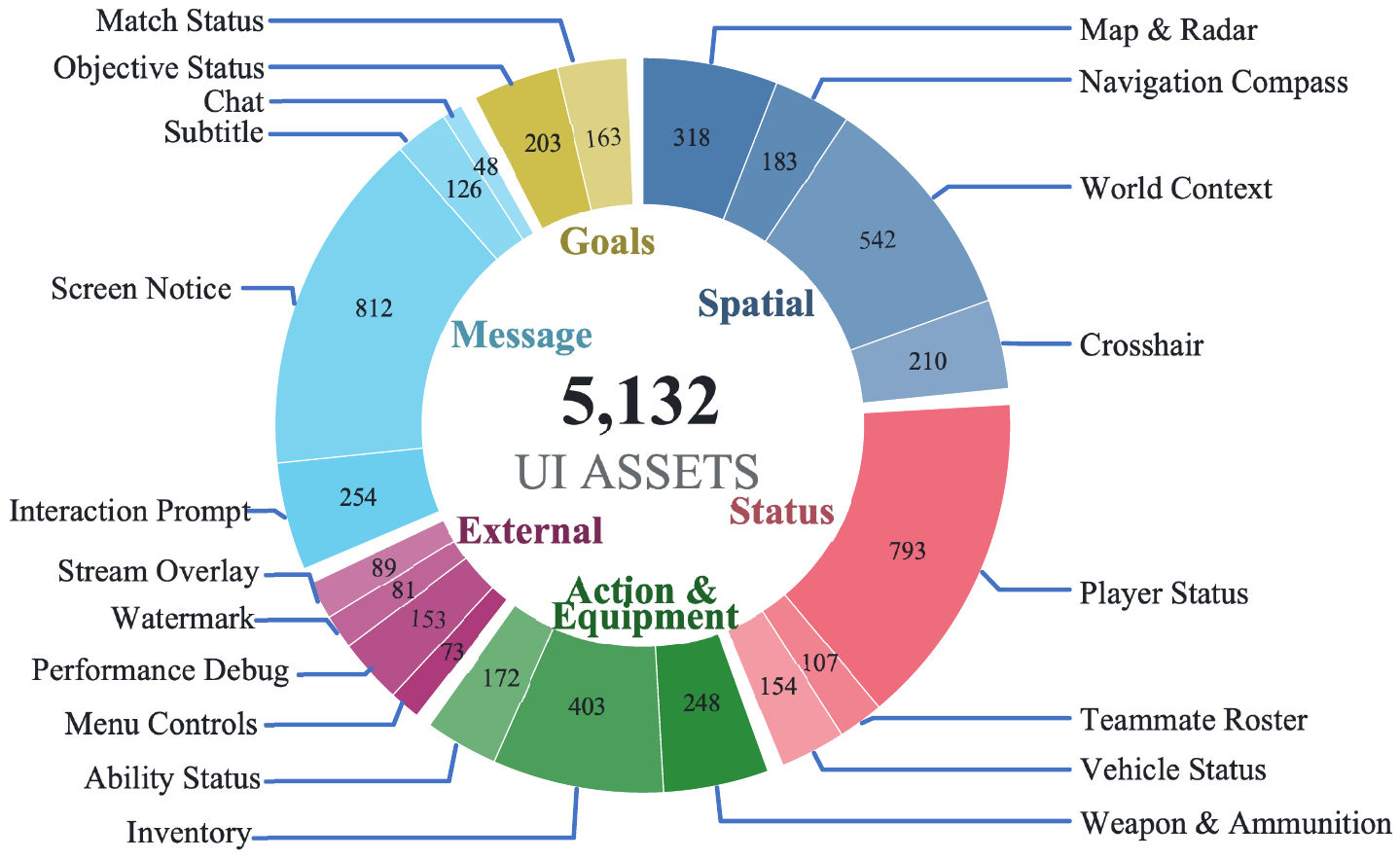}
    \caption{Statistics of the \GameDataset UI asset taxonomy.}
    \label{fig:taxonomy_stats}
\end{figure}

Our goal is to train a general gameplay UI removal model that can identify and remove diverse HUDs, menus, notifications, watermarks, and streaming overlays while faithfully preserving the underlying game content and temporal dynamics. Unlike conventional video object removal, gameplay UI removal requires semantic understanding of which visual elements belong to the game world and which are screen-space overlays. We therefore design the model to perform mask-free removal directly from the source video, enabling it to generalize across games with substantially different interface styles, layouts, and temporal behaviors. 

\begin{wrapfigure}{r}{0.45\textwidth}
    \centering
    \vspace{-10pt}
    \includegraphics[width=\linewidth]{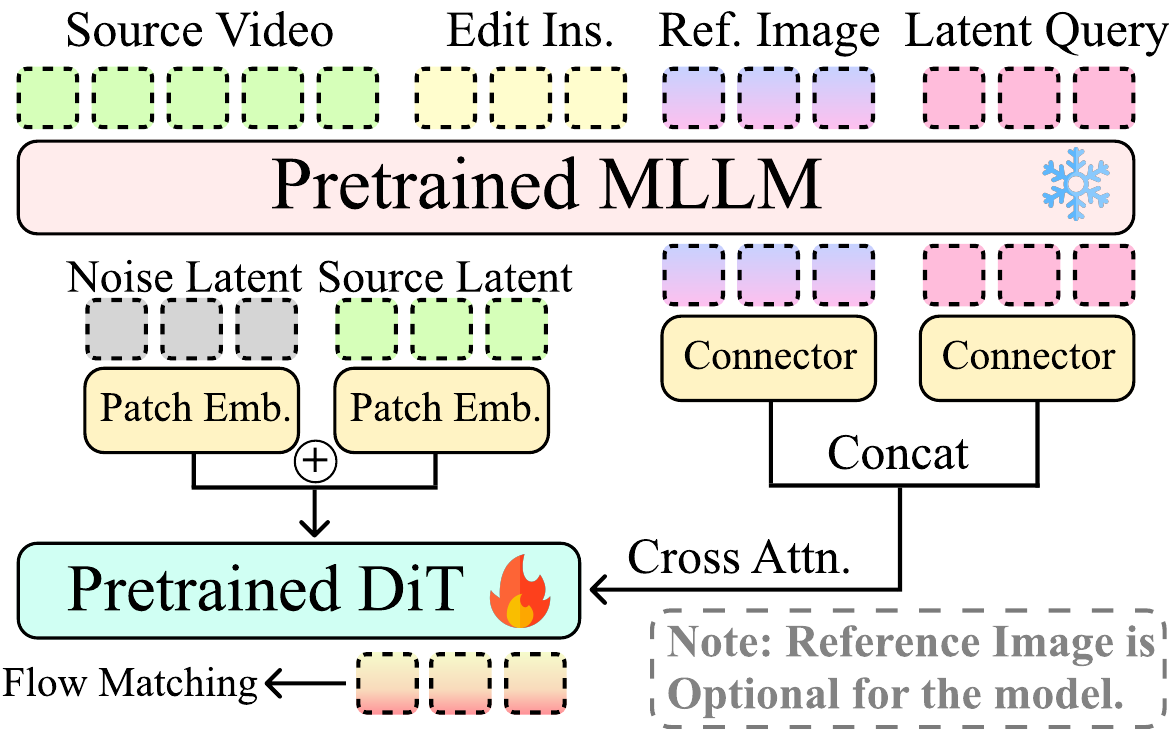}
    \caption{Model architecture of our \method.}
    \label{fig:model_arc}
    \vspace{-10pt}
\end{wrapfigure}
\paragraph{Model Architecture.} Our architecture follows Kiwi-Edit~\citep{kiwiedit2026}, coupling an MLLM encoder with a video diffusion transformer (DiT), as shown in Fig. \ref{fig:model_arc}. Given a UI-overlaid source video $\mathbf{x}_{\mathrm{src}}$ and a removal instruction $\mathbf{y}$, the MLLM jointly encodes the sampled source frames and textual instruction. A set of learnable latent queries $\mathbf{Q}$ extracts task-relevant semantic information from the MLLM features, which is projected into the DiT conditioning space: 

\begin{equation} \mathbf{c} = \mathcal{P} \left( \operatorname{CrossAttn} \left( \mathbf{Q}, \mathcal{E}_{\mathrm{MLLM}} (\mathbf{x}_{\mathrm{src}},\mathbf{y}) \right) \right), 
\end{equation}

where $\mathcal{P}$ denotes the query connector. Replacing the conventional T5 text encoder with an MLLM is important because UI removal requires visual-semantic recognition rather than text understanding alone: the MLLM can recognize and localize interface elements from their appearance and context, providing the general HUD understanding needed to transfer across unseen games. To preserve the spatial and temporal structure of the source video, its VAE latents are further injected into the noisy target latents through a timestep-dependent residual connection: 

\begin{equation} \mathbf{h}_{t} = \operatorname{PE}(\mathbf{z}_{t}) + \gamma(t)\, \operatorname{PE}_{\mathrm{src}} \left( \operatorname{VAE}(\mathbf{x}_{\mathrm{src}}) \right), \end{equation} 
where $\operatorname{PE}$ is patch embedding layer and $\gamma(t)$ is a learnable timestep-dependent scalar. 

\paragraph{Training.} We initialize the model from the Stage-2 and Stage-3 (for \emph{w.o.} and \emph{w.} reference pretraining seperately) instruction-editing checkpoint of Kiwi-Edit and post-train it on our paired UI-overlaid and clean gameplay videos at $720$p resolution. We apply LoRA to the DiT and optimize the model using the standard flow-matching objective: 
\begin{equation} \mathcal{L}_{\mathrm{flow}} = \mathbb{E}_{t,\mathbf{z}_{0},\mathbf{z}_{1},\mathbf{c}} \left[ \left\| \mathbf{v}_{\theta}(\mathbf{z}_{t},t,\mathbf{c}) - (\mathbf{z}_{1}-\mathbf{z}_{0}) \right\|_{2}^{2} \right], 
\end{equation} 
where $\mathbf{z}_{1}$ is the latent representation of the clean target video, $\mathbf{z}_{0}$ is Gaussian noise, and $\mathbf{c}$ contains the MLLM-derived removal condition. This formulation allows the model to learn both semantic identification of gameplay UI and temporally coherent reconstruction of the occluded game world.

%% file: Sec/7-experiment.tex
\section{Experiments}

\subsection{Experiment Setups}

\paragraph{Baseline Models.} Based on \textsc{Game2World}, we evaluate a diverse set of recent general video editing and object removal models for UI removal, including Aurora~\citep{yu2026aurora}, Kiwi-Edit~\citep{lin2026kiwiedit}, LoomVideo~\citep{wu2026loomvideo}, LTX-2~\citep{hacohen2026ltx2}, Lucy-Edit-5B~\citep{decart2025lucyedit}, OmniWeaving~\citep{pan2026omniweaving}, VACE-14B~\citep{vace}, and EffectErase~\citep{fu2026EffectErase}. 

\paragraph{Model Details.}
We initialize our model from the Stage-2 (w.o. reference pretrain) and Stage-3 (w. reference pretrain) checkpoint of Kiwi-Edit~\citep{kiwiedit2026}, leveraging its general visual-semantic understanding of UI elements in gameplay videos. Fot Stage-3 checkpoint training, we random drop 20\% of reference clean images to enable model's reference-free UI removal capability. During evalution, we provide both results of w. and w.o. reference images. We freeze the MLLM encoder and connector, apply LoRA with rank 64~\citep{hu2021lora} to the DiT module, and fine-tune the model on \GameDatasetS using  8*H100 GPUs with a global batch size of 32, a learning rate of $1\times10^{-4}$, and 3,000 training steps.

\begin{figure*}[!t]
    \centering
    \includegraphics[width=\linewidth]{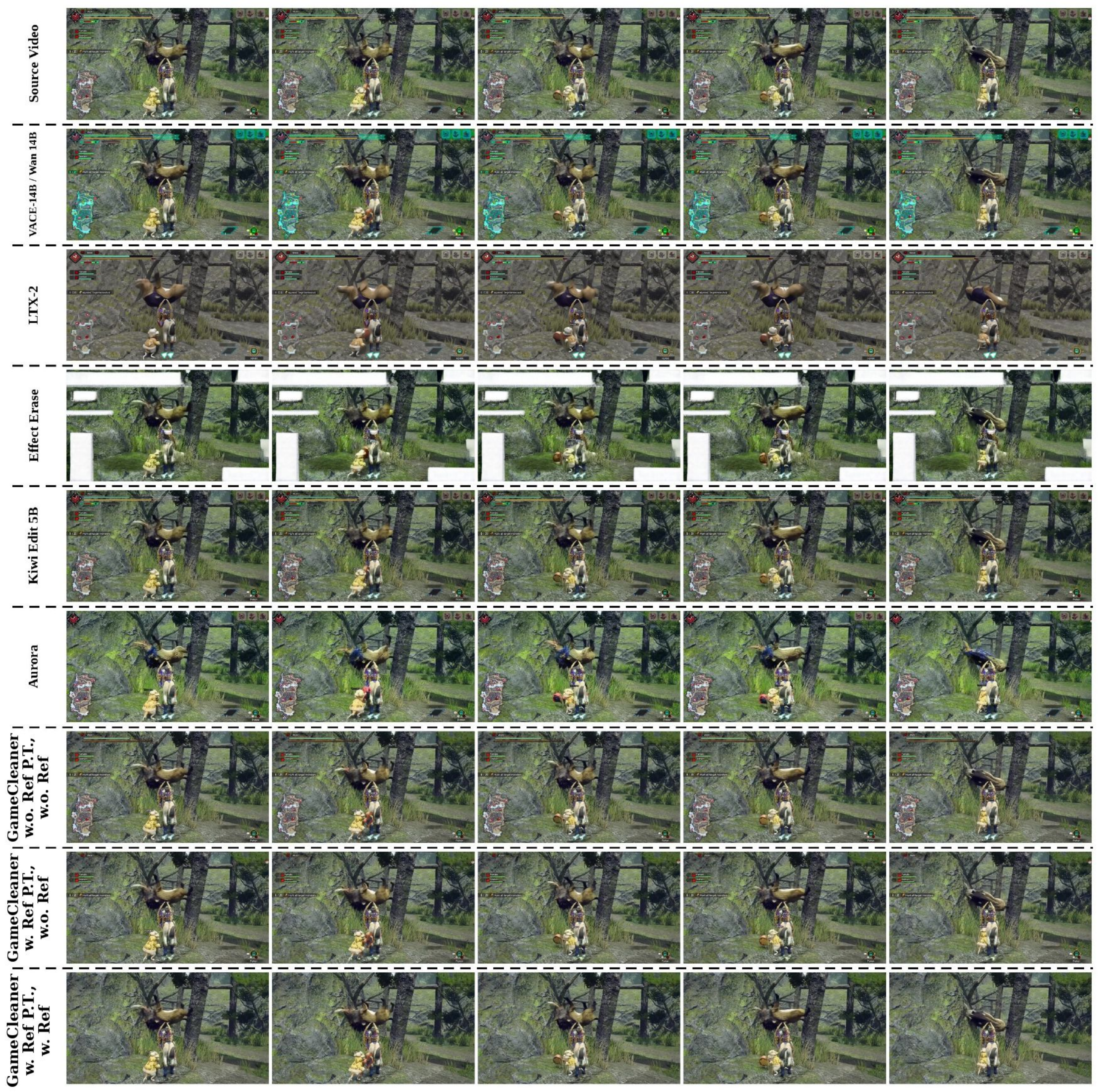}
    \caption{Qualitative study for UI removal. \method w. reference image substantially reduces UI overlay while faithfully retaining the original video content. Frame and box level evaluation case study are provided in the Appendix.}
    \label{fig:qualitative_study}
\end{figure*}

\paragraph{Evaluation Protocol.}
We sample all videos at 2 FPS and use an MLLM judge for evaluation. For the Synthetic benchmark, the judge compares the restored frame with its UI-free ground truth. For the In-the-Wild benchmark, it receives the original frame, the frame with annotated UI bounding boxes, and the restored output. We resize the raw and edited keyframes into 1280*720p for all models for fair evaluation. We report four metrics:

\begin{itemize}[leftmargin=*,itemsep=0pt]
    \item \textbf{Artificial-Adjusted Removal (AAR):} Removed elements receive scores of $1.0$, $0.5$, or $0.0$ for no, minor, or major artifacts, respectively; all unsuccessful or uncertain removals receive $0$.
    
    \item \textbf{UI:} Each annotated element is labeled as \texttt{removed}, \texttt{present}, or \texttt{uncertain}, receiving a score of $1$ only when successfully removed, regardless of restoration artifacts.
    
    \item \textbf{Clean:} An element receives a score of $1$ only when it is removed without any visible artifact; minor blur, seams, or texture defects are counted as failures.
    
    \item \textbf{Background Preservation (BG):} Preservation outside annotated UI regions receive $1.0$, $0.5$, and $0.0$ scores for unchanged, minor-change, and major-change, respectively.

    \item \textbf{Uncertain: Evaluation uncertainty (confidence) provided by the judge model.}
\end{itemize}

% For the first three metrics, scores are macro-averaged from elements to frames, frames to videos, and videos to each benchmark. Background Preservation is averaged from frames to videos and then across videos.
We validate the MLLM judge against human annotations on 200 videos, covering 6150 UI-element labels and 2000 frame-level labels. It achieves a removed F1 of 87.78\%, artifact agreement of 85.96\%, preservation agreement of 78.00\%, and Cohen's $\kappa$ of 0.820, demonstrating strong alignment with human evaluation. We also evaluate the inter-rater reliability in Table \ref{tab:inter_rater_reliability}.

% ============================================================
% Main Results
% ============================================================

\begin{table*}[t]
\centering
\caption{
Main comparison on the synthetic and in-the-wild benchmarks.
\textit{Ref P.T.} denotes reference-based post-training.
Overall is computed as
$(\mathrm{AAR}_{S}+\mathrm{BG}_{S}+\mathrm{AAR}_{W}+\mathrm{BG}_{W})/4$,
where $S$ and $W$ denote the synthetic and wild benchmarks.
}
\label{tab:main_hud_removal_results}
\setlength{\tabcolsep}{3.2pt}
\renewcommand{\arraystretch}{1.12}

\resizebox{\textwidth}{!}{
\begin{tabular}{lccc ccccc ccccc}
\toprule[1.5pt]

\textbf{Model}
&
\textbf{Mask}
&
\textbf{Ref P.T.}
&
\textbf{Overall}
&
\multicolumn{5}{c}{\textbf{Synthetic}}
&
\multicolumn{5}{c}{\textbf{In-the-Wild}}
\\

\cmidrule(lr){5-9}
\cmidrule(lr){10-14}

&
&
&
&
\textbf{AAR}
&
\textbf{UI}
&
\textbf{Clean}
&
\textbf{BG}
&
\textbf{Uncertain $\downarrow$}
&
\textbf{AAR}
&
\textbf{UI}
&
\textbf{Clean}
&
\textbf{BG}
&
\textbf{Uncertain $\downarrow$}
\\

\midrule

VACE 1.3B \citep{vace}& \ding{52} & \ding{56} & 69.37 & 59.11 & 72.29 & 52.14 & 96.50 & 0.26 & 30.40 & 38.28 & 27.13 & 91.45 & 0.29 \\
EffectErase \citep{fu2026EffectErase}& \ding{52} & \ding{56} & 59.68 & 56.17 & 94.56 & 48.75 & 84.45 & 0.07 & 23.55 & \textbf{92.00} & 20.42 & 74.55 & 0.09 \\
VACE 14B \citep{vace}& \ding{52} & \ding{56} & 56.87 & 27.86 & 33.12 & 27.12 & \textbf{99.60} & 0.37 & 5.92 & 7.68 & 5.37 & 94.10 & 0.11 \\

\midrule

Aurora \citep{yu2026aurora}& \ding{56} & \ding{56} & 37.57 & 60.62 & 66.68 & 59.14 & 24.45 & 0.20 & 51.19 & 55.87 & 49.41 & 14.00 & 0.16 \\
LTX-2 \citep{hacohen2026ltx2}& \ding{56} & \ding{56} & 17.21 & 43.25 & 50.23 & 43.00 & 2.25 & 0.38 & 20.62 & 33.02 & 19.16 & 2.70 & 0.36 \\
LoomVideo \cite{wu2026loomvideo}& \ding{56} & \ding{56} & 42.82 & 31.49 & 35.03 & 30.06 & 87.50 & 0.20 & 30.35 & 35.92 & 28.11 & 21.95 & 0.38 \\
Kiwi-Edit \citep{kiwiedit2026}& \ding{56} & \ding{56} & 60.58 & 34.68 & 40.97 & 31.28 & 97.30 & 0.11 & 13.54 & 15.70 & 13.01 & 96.80 & 0.40 \\
OmniWeaving \citep{pan2026omniweaving}& \ding{56} & \ding{56} & 35.31 & 0.95 & 1.55 & 0.54 & 86.70 & 0.23 & 5.27 & 5.74 & 4.88 & 48.30 & 0.20 \\
Lucy-Edit \citep{decart2025lucyedit}& \ding{56} & \ding{56} & 42.62 & 1.30 & 1.36 & 1.26 & 83.30 & 0.42 & 2.86 & 3.40 & 2.54 & 83.00 & 0.25 \\

\midrule

\textbf{\method (w. Ref)} & \ding{56} & \ding{52} & \textbf{93.34} & \underline{94.64} & \underline{97.56} & \underline{92.75} & 98.85 & \textbf{0.04} & \textbf{80.05} & \underline{84.75} & \textbf{77.84} & \textbf{99.80} & \underline{0.07} \\
\textbf{\method (w.o. Ref)} & \ding{56} & \ding{52} & \underline{85.97} & 93.17 & \textbf{97.78} & 89.17 & \underline{99.50} & \underline{0.05} & \underline{51.60} & 55.55 & \underline{49.43} & \underline{99.60} & 0.09 \\
\textbf{\method} & \ding{56} & \ding{56} & 85.63 & \textbf{95.36} & 97.43 & \textbf{93.59} & 99.00 & \underline{0.05} & 48.56 & 49.85 & 47.65 & \underline{99.60} & \textbf{0.02} \\

\bottomrule[1.5pt]
\end{tabular}
}
\end{table*}

% ============================================================
% Fine-Grained Results
% ============================================================

\begin{table*}[t]
\centering
\caption{
Fine-grained UI Removal results on the synthetic and
in-the-wild benchmarks.
\textit{Ref P.T.} means reference based post train.
}
\label{tab:fine_grained_hud_removal_results}
\setlength{\tabcolsep}{4.5pt}
\renewcommand{\arraystretch}{1.12}

\resizebox{\textwidth}{!}{
\begin{tabular}{lcc cccc cccc}
\toprule[1.5pt]

\textbf{Model}
&
\textbf{Mask}
&
\textbf{Ref P.T.}
&
\multicolumn{4}{c}{\textbf{Synthetic}}
&
\multicolumn{4}{c}{\textbf{In-the-Wild}}
\\

\cmidrule(lr){4-7}
\cmidrule(lr){8-11}

&
&
&
\textbf{Frame-Macro}
&
\textbf{Box-Micro}
&
\textbf{Track-All}
&
\textbf{Track-Clean}
&
\textbf{Frame-Macro}
&
\textbf{Box-Micro}
&
\textbf{Track-All}
&
\textbf{Track-Clean}
\\

\midrule

VACE 1.3B \citep{vace}& \ding{52} & \ding{56} & 72.26 & 75.29 & 74.58 & 56.09 & 37.33 & 31.69 & 33.88 & 20.29 \\
EffectErase \citep{fu2026EffectErase}& \ding{52} & \ding{56} & 94.55 & 95.00 & 94.87 & 41.77 & \textbf{91.85} & \textbf{93.29} & \textbf{92.93} & 16.12 \\
VACE 14B \citep{vace}& \ding{52} & \ding{56} & 33.10 & 32.41 & 31.74 & 26.37 & 6.57 & 3.79 & 5.25 & 4.35 \\

\midrule

Aurora \citep{yu2026aurora}& \ding{56} & \ding{56} & 66.66 & 67.21 & 65.87 & 56.80 & 56.09 & 52.67 & 53.44 & \underline{56.80} \\
LTX-2 \citep{hacohen2026ltx2}& \ding{56} & \ding{56} & 50.28 & 61.74 & 63.13 & 54.65 & 32.01 & 30.61 & 31.34 & 15.76 \\
LoomVideo \citep{wu2026loomvideo}& \ding{56} & \ding{56} & 34.97 & 29.86 & 30.79 & 26.01 & 34.86 & 27.54 & 27.90 & 23.37 \\
Kiwi-Edit \citep{kiwiedit2026}& \ding{56} & \ding{56} & 40.92 & 43.59 & 40.57 & 32.34 & 15.26 & 12.14 & 11.23 & 9.06 \\
OmniWeaving \citep{pan2026omniweaving}& \ding{56} & \ding{56} & 1.55 & 1.21 & 1.07 & 0.36 & 5.47 & 4.31 & 3.62 & 3.26 \\
Lucy-Edit \citep{decart2025lucyedit}& \ding{56} & \ding{56} & 1.36 & 0.93 & 0.84 & 0.60 & 3.50 & 2.45 & 1.81 & 1.27 \\

\midrule

\textbf{\method (w. Ref)} & \ding{56} & \ding{52} & \underline{97.56} & \textbf{98.72} & \textbf{98.57} & \textbf{95.47} & \underline{85.34} & \underline{83.91} & \underline{76.81} & \textbf{68.12} \\
\textbf{\method (w.o. Ref)} & \ding{56} & \ding{52} & \textbf{97.78} & 98.07 & 97.61 & 92.00 & 56.64 & 52.40 & 46.56 & 41.85 \\
\textbf{\method} & \ding{56} & \ding{56} & 97.43 & \underline{98.26} & \underline{97.85} & \underline{94.75} & 49.20 & 45.20 & 41.49 & 38.22 \\

\bottomrule[1.5pt]
\end{tabular}
}
\end{table*}

\begin{figure*}[!t]
    \centering
    \includegraphics[width=\linewidth]{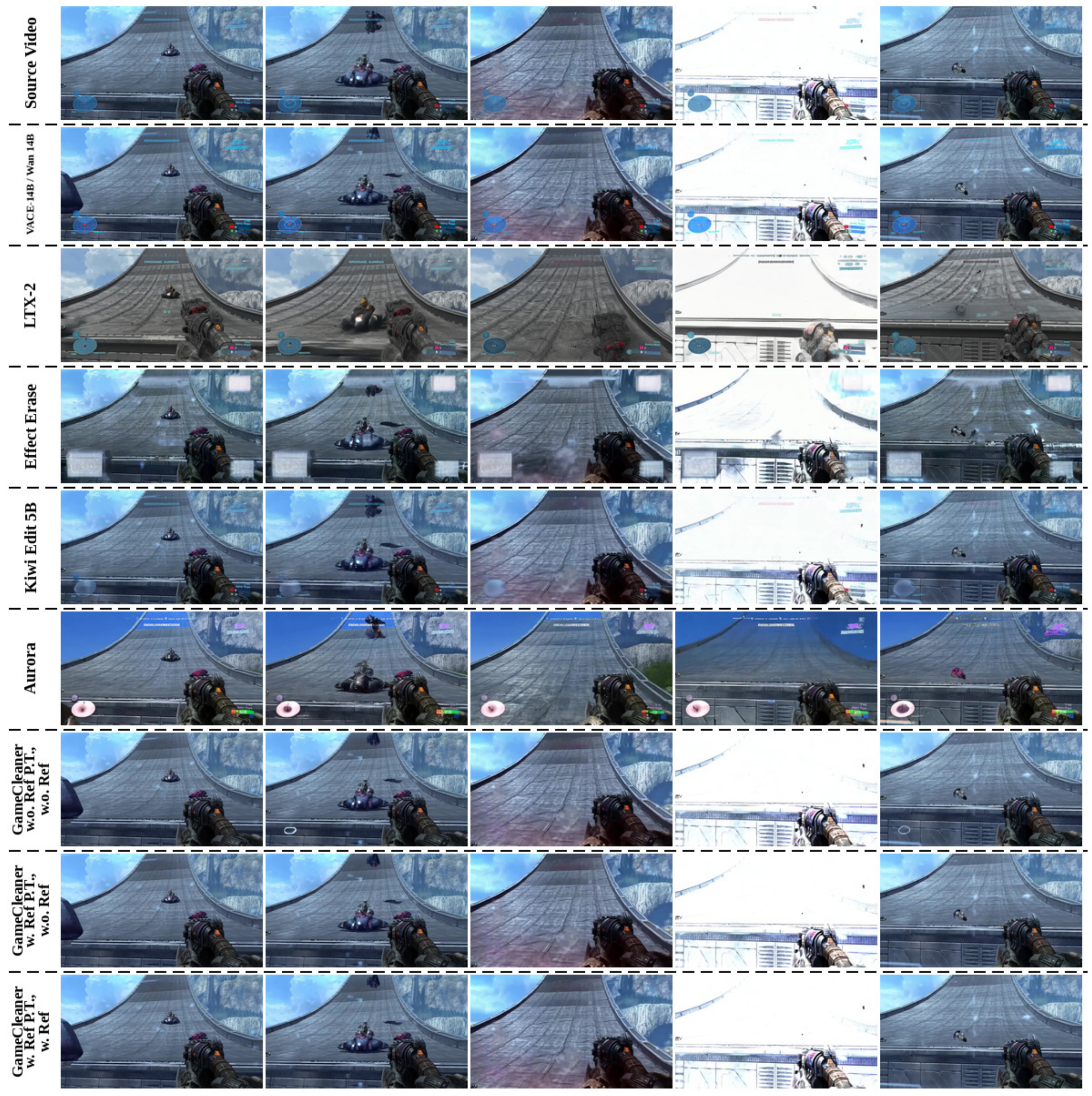}
    \caption{Qualitative study of the in-the-wild subset.}
    \label{fig:qualitative_study_wild2}
\end{figure*}

\begin{figure*}[!t]
    \centering
    \includegraphics[width=\linewidth]{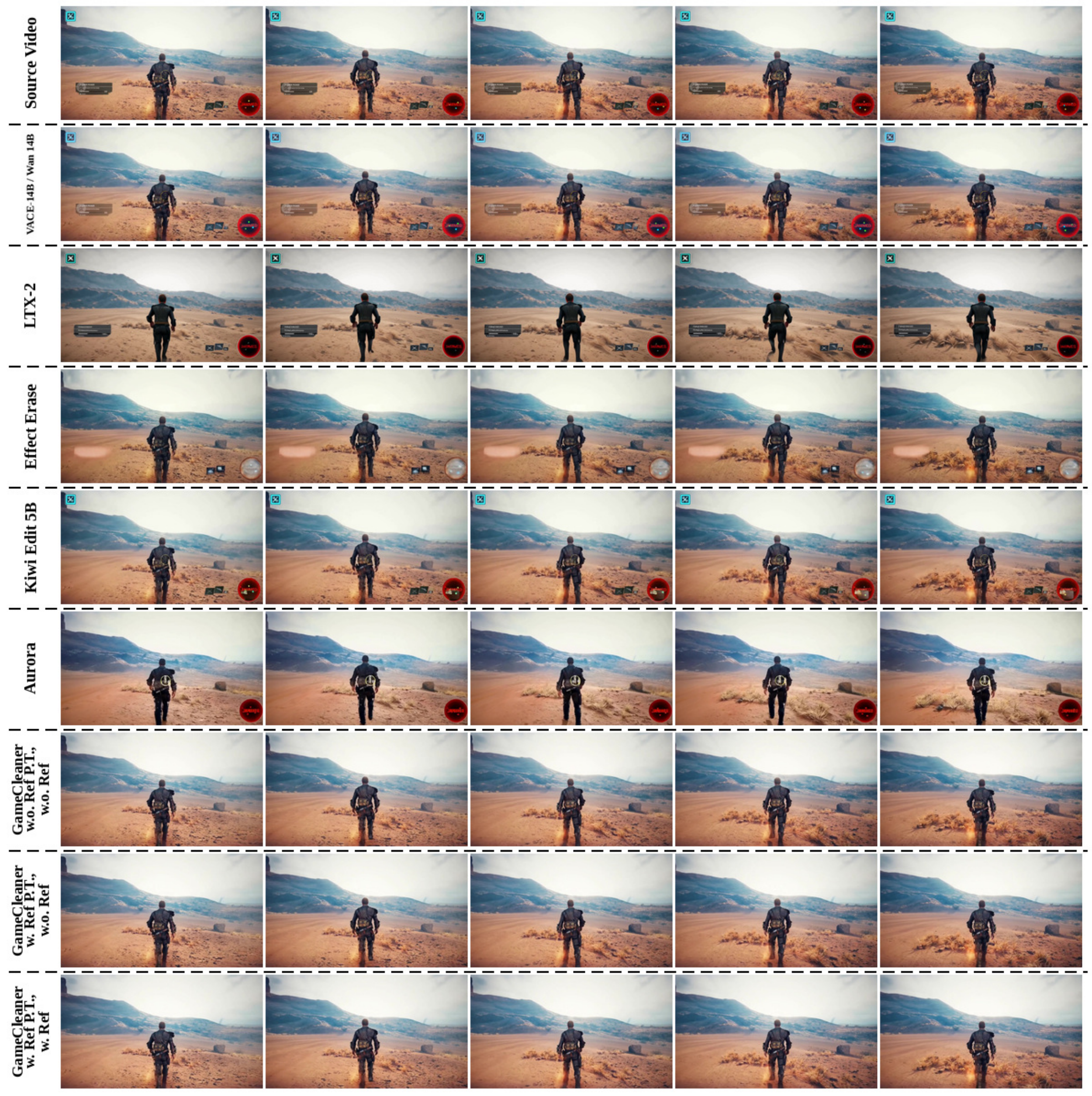}
    \caption{Qualitative study of the synthetic subset.}
    \label{fig:qualitative_study_syn}
\end{figure*}

\subsection{Experimental Results}

\begin{figure*}[!t]
    \centering
    \includegraphics[width=\linewidth]{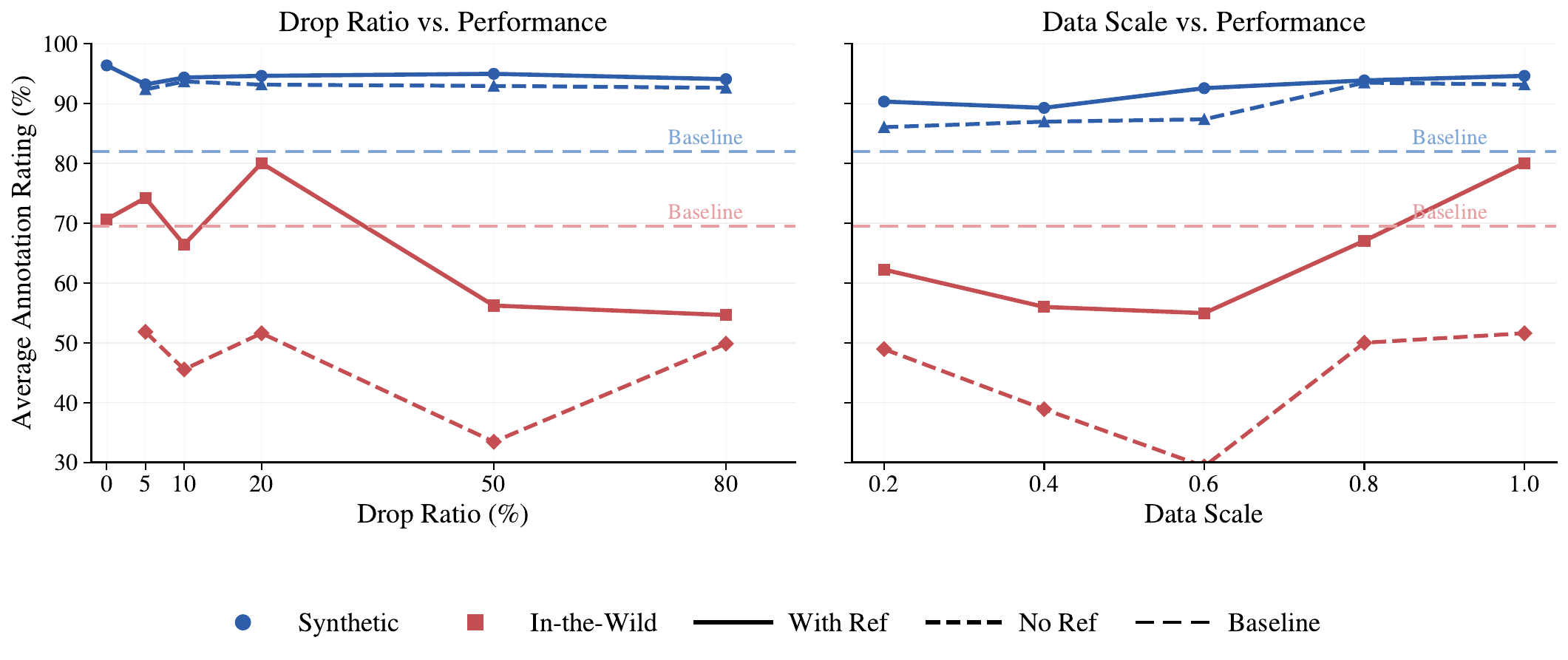}
    \caption{Ablation study of data scaling and reference drop ratio during training. We find out that larger and more diverse dataset show generalizable result on in-the-wild evaluation and have not converge yet.}
    \label{fig:qualitative_study_wild1}
\end{figure*}

\paragraph{Overall Performance.}
As shown in Table~\ref{tab:main_hud_removal_results}, \method achieves the strongest overall removal--preservation trade-off. The reference-conditioned variant obtains an Overall score of 93.34, outperforming the strongest baseline, mask-assisted VACE-1.3B, by 23.97 percentage points and the strongest mask-free baseline, Kiwi-Edit, by 32.76 points. This advantage does not depend entirely on providing a reference at inference time: \method (w.o. Ref) and the standard \method achieve 85.97 and 85.63, respectively. Since Overall averages AAR and BG across both benchmarks, these improvements indicate that \method removes UI elements effectively without achieving higher removal scores by unnecessarily modifying the underlying game world.

\paragraph{Synthetic Benchmark.}
On \GameDataset-S, all three variants of \method establish a strong balance between artifact-adjusted removal and background preservation. The standard \method achieves the highest AAR of 95.36 while retaining a BG score of 99.00, whereas the reference-conditioned and reference-free post-trained variants obtain AAR/BG scores of 94.64/98.85 and 93.17/99.50, respectively. In comparison, Aurora reaches only 60.62 AAR with 24.45 BG, while VACE-14B achieves 99.60 BG but only 27.86 AAR. EffectErase obtains a high raw UI-removal score of 94.56, yet its AAR decreases to 56.17 and its BG score to 84.45. These results demonstrate that raw removal success alone does not adequately capture the quality of the edited video.

\paragraph{In-the-Wild Benchmark.}
The advantage of \method becomes more pronounced on \GameDataset-W. The reference-conditioned variant achieves 80.05 AAR and 99.80 BG, exceeding the strongest baseline AAR, obtained by Aurora, by 28.86 percentage points while improving its BG score by 85.80 points. Although EffectErase obtains a raw UI-removal score of 92.00, its Clean score, AAR, and BG decrease to 20.42, 23.55, and 74.55, respectively. By contrast, \method (w. Ref) reaches 77.84 Clean removal together with near-perfect background preservation. Even without a reference at inference time, \method obtains 51.60 AAR and 99.60 BG, matching the strongest baseline in removal performance while preserving substantially more scene content.

\paragraph{Fine-Grained Removal Quality.}
Table~\ref{tab:fine_grained_hud_removal_results} further explains the differences captured by AAR. On the synthetic benchmark, \method (w. Ref) achieves 98.72 Box-Micro, 98.57 Track-All, and 95.47 Track-Clean, indicating consistent removal across individual UI regions and complete tracks. On the wild benchmark, EffectErase produces higher raw Frame-Macro, Box-Micro, and Track-All scores, but its Track-Clean score is only 16.12. In contrast, \method (w. Ref) achieves 68.12 Track-Clean despite lower raw removal scores. Thus, many of the regions modified by EffectErase contain visible artifacts or unintended scene changes, whereas \method more frequently completes the entire removal track cleanly. This result supports using artifact-adjusted metrics rather than raw detection-based removal rates alone.

\paragraph{Judge Uncertainty.}
The uncertainty metric measures how confidently the judge model can determine the evaluation outcome, with lower values indicating less ambiguous generations. The reference-conditioned variant obtains the lowest uncertainty on the synthetic benchmark at 0.04, while the standard \method obtains the lowest in-the-wild uncertainty at 0.02. The remaining \method variants also remain low, ranging from 0.05 to 0.09, whereas many baselines exhibit substantially higher uncertainty, reaching up to 0.42. Low uncertainty is not itself evidence of successful removal; however, when considered together with high AAR and BG scores, it indicates that the improvements of \method are visually clear and can be identified consistently by the judge.

\begin{wraptable}{r}{0.4\textwidth}
\centering
\vspace{-1em}
\caption{Inter-rater reliability on the in-the-wild benchmark.}
\label{tab:inter_rater_reliability}
\setlength{\tabcolsep}{4.0pt}
\renewcommand{\arraystretch}{1.12}
\resizebox{0.4\textwidth}{!}{
\begin{tabular}{lcccc}
\toprule
\textbf{Metric} & \textbf{ICC(A,1)} & \textbf{ICC(A,5)} & \textbf{Kripp. $\alpha$} & \textbf{Kendall $W$} \\
\midrule
\textbf{AAR} & 0.713 & 0.926 & 0.712 & 0.772 \\
\textbf{UI} & 0.724 & 0.929 & 0.722 & 0.777 \\
\textbf{Clean} & 0.681 & 0.914 & 0.679 & 0.748 \\
\textbf{BG} & 0.834 & 0.962 & 0.833 & 0.566 \\
\textbf{Unc. $\downarrow$} & 0.582 & 0.874 & 0.580 & 0.705 \\
\bottomrule
\end{tabular}}
\vspace{-1em}
\end{wraptable}
\paragraph{Reliability of MLLM-as-a-Judge.}
The inter-rater analysis in Table~\ref{tab:inter_rater_reliability} demonstrates that the proposed evaluation is reliable when ratings are aggregated. AAR achieves an ICC(A,1) of 0.713, an ICC(A,5) of 0.926, a Krippendorff's $\alpha$ of 0.712, and a Kendall's $W$ of 0.772, showing good absolute agreement and ranking consistency. UI and Clean exhibit similarly high aggregated ICC values of 0.929 and 0.914, respectively. BG obtains the highest absolute agreement, with ICC(A,1)/ICC(A,5) of 0.834/0.962 and $\alpha=0.833$, although its rank agreement is more moderate at $W=0.566$. Uncertainty is less stable as an individual rating, with ICC(A,1) of 0.582, but reaches 0.874 after aggregating five ratings. These results justify using aggregated judge scores while also quantifying the residual variability of each metric.

\paragraph{Ablation Study.}
The reference-drop ablation identifies a 20\% drop ratio as the most effective balance between reference-conditioned performance and reference-free robustness. With reference inference, increasing the drop ratio from 0\% to 20\% improves wild AAR from 70.64 to 80.05, while synthetic AAR remains high at 94.64 compared with 96.39. More aggressive dropping is detrimental: ratios of 50\% and 80\% reduce wild AAR to 56.23 and 54.64. Under reference-free inference, the 20\% setting achieves 93.17/51.60 AAR on the synthetic/wild benchmarks, within 0.24 points of the best reference-free wild result. Data scaling exhibits a complementary pattern. From 60\% to 100\% of training, reference-conditioned synthetic AAR increases by only 2.06 points, from 92.58 to 94.64, whereas wild AAR rises by 25.09 points, from 54.96 to 80.05. The corresponding reference-free gains are 5.80 and 22.28 points. Thus, synthetic performance begins to saturate earlier, while additional data produces substantially larger improvements in real-world generalization. The non-monotonic intermediate results further suggest that sufficient data diversity, rather than training progress alone, is necessary to prevent overfitting to synthetic HUD compositions.

\begin{wraptable}{r}{0.5\textwidth}
\centering
\vspace{-1em}
\caption{Effect of inference resolution on the synthetic and in-the-wild benchmarks.}
\label{tab:resolution_ablation}
\setlength{\tabcolsep}{3.2pt}
\renewcommand{\arraystretch}{1.08}
\resizebox{0.5\textwidth}{!}{
\begin{tabular}{lc ccc ccc}
\toprule
\textbf{Set} & \textbf{Resolution} & \multicolumn{3}{c}{\textbf{Synthetic}} & \multicolumn{3}{c}{\textbf{In-the-Wild}} \\
\cmidrule(lr){3-5}\cmidrule(lr){6-8}
& & \textbf{AAR} & \textbf{BG} & \textbf{Uncertain $\downarrow$} & \textbf{AAR} & \textbf{BG} & \textbf{Uncertain $\downarrow$} \\
\midrule
\multirow{3}{*}{Set 1} & $800{\times}448$ & 31.03 & 95.15 & 0.24 & 18.27 & 97.42 & 0.19 \\
& $1280{\times}720$ & 28.12 & 97.47 & 0.46 & 16.43 & 98.35 & 0.24 \\
& $1920{\times}1080$ & 32.02 & 95.76 & 0.44 & 16.20 & 95.46 & 0.09 \\
\midrule
\multirow{3}{*}{Set 2} & $800{\times}448$ & 86.19 & 97.47 & 0.03 & 33.47 & 99.23 & 0.09 \\
& $1280{\times}720$ & 86.08 & 99.04 & 0.22 & 38.49 & 99.79 & 0.19 \\
& $1920{\times}1080$ & 87.95 & 97.98 & 0.04 & 37.48 & 98.92 & 0.45 \\
\bottomrule
\end{tabular}
}
\vspace{-1em}
\end{wraptable}
\paragraph{Effect of Judge Resolution.}
The resolution study shows that judge scores remain sensitive to the spatial resolution of the evaluated videos. This effect appears in both Set~1 and Set~2, even though the two sets contain outputs from different mixtures of generation models, indicating that evaluation resolution should be standardized and explicitly reported. All resolution variants were evaluated using the same judging protocol and model endpoint within a single day. Consequently, the observed differences are unlikely to be caused by endpoint updates or provider-side model drift, although variation between the two sets can still reflect their different model compositions. These findings motivate retaining uncertainty and multi-rating reliability measurements alongside the primary quality scores.

\paragraph{Qualitative Results.}
As shown in Fig.~\ref{fig:qualitative_study_wild1}, \ref{fig:qualitative_study_wild2}, \ref{fig:qualitative_study_syn}, \method removes diverse static and temporally varying HUD elements while preserving scene geometry, textures, character appearance, and temporal continuity. EffectErase frequently removes the target UI but introduces changes in unrelated regions, whereas VACE-14B tends to preserve the background at the cost of visible UI residuals. These examples are consistent with the quantitative AAR, BG, and Track-Clean results and demonstrate the suitability of \method for converting in-the-wild gameplay videos into cleaner world-model training data.

%% file: Sec/8-discussion.tex
\section{Discussion and Limitations}

\paragraph{Scaling Behavior and Task Scalability.}
\engine is built on the premise that gameplay UI removal should be treated as a scalable data-processing problem rather than a game-specific editing task. Although our current training study is limited to approximately 100K paired videos, the performance of \method consistently improves as the amount of training data increases, without showing clear saturation within the evaluated range. This trend suggests that interface--world disentanglement benefits from broader coverage of game genres, visual styles, UI layouts, and temporal behaviors, and that gameplay UI removal is amenable to further data scaling. Nevertheless, our results should be interpreted as evidence of a favorable scaling trend rather than a definitive power-law relationship. Establishing a formal scaling law will require experiments spanning substantially larger data, model, and compute scales. We are currently scaling the dataset to 1M scale from more diverse sources.

\paragraph{Scalable Data Collection and Processing.}
In-the-wild gameplay videos provide an abundant and continuously expanding source of visual interaction data. \engine makes this data practically usable by converting heterogeneous gameplay footage into structured supervision and cleaned videos through a unified pipeline for UI asset extraction, clean-video curation, temporally coherent UI synthesis, and mask-free UI removal. Most stages operate automatically at the clip level and can be parallelized, allowing the processed dataset to grow together with the available raw footage. The resulting data retains diverse game-world appearances and dynamics while reducing game-specific screen-space artifacts, making it more suitable for video-based world-model training. This scalability distinguishes our approach from data pipelines that depend on game-specific instrumentation, privileged simulator states, or manually specified masks.

\paragraph{Limitations and Future Directions.}
Our current work focuses on establishing the data engine, task formulation, benchmark, and a generalist UI-removal model, rather than maximizing training or inference scale. Consequently, \GameDataset does not yet cover the full diversity of games, interfaces, and recording conditions, and \method may produce imperfect reconstructions for large opaque interfaces, rapidly changing overlays, or UI elements visually entangled with the underlying scene. Moreover, our downstream study uses text-conditioned video generation as a controlled proxy for world-model training rather than a fully action-conditioned gameplay world model. While cleaned gameplay videos provide observations of world dynamics, interactive world-model training additionally requires recovering player actions, for which broadly reliable inverse-dynamics or latent-action models across heterogeneous games and action spaces remain unavailable.

Our future work will therefore proceed along three directions: \textbf{\textit{(i) Data and benchmark scaling.}} We will expand \GameDataset to substantially more games, UI types, and recording conditions, and construct broader and more challenging benchmarks. \textbf{\textit{(ii) Action-conditioned world modeling.}} We will develop general action-recovery modules to convert cleaned gameplay videos into observation--action trajectories and directly train gameplay world models on them, providing stronger end-to-end validation of the processed data. \textbf{\textit{(iii) Efficient \method.}} An ideal \method should operate as a few-step distilled model rather than requiring the full generation trajectory used in our current pipeline. Because few-step distillation requires substantial additional compute and careful ablations, we leave it to future work together with other inference acceleration and GPU-level optimization techniques; the present work instead isolates and studies the direct data-processing and model-training pipeline.

%% file: Sec/10-conclusion.tex
\section{Conclusion}
We formulate gameplay UI removal as interface--world disentanglement and introduce a unified taxonomy, a scalable data engine, a large-scale benchmark, and a mask-free removal model. Our experiments demonstrate that removing gameplay UI can preserve underlying scene content and temporal dynamics while improving the value of gameplay videos for downstream world-model training. We hope this work provides an essential step toward unlocking large-scale in-the-wild gameplay data for future world models.

\section*{Ackownledgement}
We thank Yaochen Wang, Chenlong Wang, and Prof. Tianyi Zhou for their invaluable support and discussion. Dongping Chen is supported by a Modal Academic Grant.

%% file: Sec/99-appendix.tex
% -----------------------------------------------------------------------------
% Game2World supplementary appendix draft
% Verified revision: 2026-08-01-v4-method-focused
% Sources: submitted manuscript, supplied supplementary archive, and frozen UI evaluation protocol.
% Recommended preamble packages:
% \usepackage{amsmath,amssymb,booktabs,array,multirow,listings}
% \usepackage{xcolor}
% The appendix can be included with \input{appendix_game2world_v4_method_focused.tex}.
% -----------------------------------------------------------------------------

\lstdefinestyle{g2wcode}{
  basicstyle=\ttfamily\scriptsize,
  breaklines=true,
  breakatwhitespace=false,
  columns=fullflexible,
  keepspaces=true,
  showstringspaces=false,
  frame=single,
  framerule=0.3pt,
  xleftmargin=0.5em,
  xrightmargin=0.5em
}

% \section{Appendix}
% \label{app:scope}

% This appendix reports implementation, data-construction, annotation, model-training, inference, and evaluation details documented in the submitted manuscript, the supplied supplementary code package, and the frozen HUD-removal evaluation protocol.

\section{Full \taxonomy}
\label{app:taxonomy}
We distinguish different types of UI elements based on their functions and the screen regions where they commonly appear. We first selected diverse and representative screenshots from more than 500 games and performed a coarse categorization of the labels based on their functions and names. We then iteratively refined and merged these categories, eventually arriving at the current taxonomy. The taxonomy contains 21 labels, including the \texttt{other\_hud} fallback category. Table~\ref{tab:full-taxonomy} reproduces the definitions used by the supplied renderer.

\begin{table*}[!h]
\centering
\caption{Complete taxonomy definitions used for UI annotation and synthesis.}
\label{tab:full-taxonomy}
\vspace{2pt}
\scriptsize
\setlength{\tabcolsep}{4pt}
\renewcommand{\arraystretch}{1.10}
\begin{tabular}{@{}>{\raggedright\arraybackslash}p{0.35\textwidth}>{\raggedright\arraybackslash}p{0.60\textwidth}@{}}
\toprule
\textbf{Label} & \textbf{Definition}\\
\midrule
\texttt{map\_radar} & Geographic mini-map or symbolic proximity/scan radar.\\
\texttt{navigation\_compass} & Compass, heading scale, bearing strip, or directional readout.\\
\texttt{player\_status} & Health, armor, stamina, resources, level, portrait, or player-state cluster.\\
\texttt{teammate\_roster} & Party/team member list or teammate status cluster.\\
\texttt{weapon\_ammo} & Equipped weapon, ammunition, reload, or firing-mode widget.\\
\texttt{inventory} & Hotbar, quick slots, backpack, item grid, or inventory panel.\\
\texttt{ability\_status} & Skills, powers, charges, or cooldowns.\\
\texttt{vehicle\_status} & Vehicle controls, gauges, or telemetry.\\
\texttt{objective\_status} & Mission/task text, counters, or objective progress.\\
\texttt{match\_status} & Timer, score, round, placement, scoreboard, or match progress.\\
\texttt{world\_context} & World markers, names, locations, distance, opponent, or world-object state.\\
\texttt{crosshair} & Aiming reticle.\\
\texttt{interaction\_prompt} & Immediate key/button action prompt.\\
\texttt{screen\_notice} & Notification, tutorial, kill feed, pickup, warning, or achievement.\\
\texttt{subtitle} & Dialogue caption.\\
\texttt{chat\_comms} & Typed player-to-player chat or chat log.\\
\texttt{menu\_controls} & Menu, toolbar, loading controls, cursor, or utility controls.\\
\texttt{performance\_debug} & Performance, network, diagnostics, version, or debug readout.\\
\texttt{watermark} & Passive copyright, publisher, sponsor, recording, channel, logo, or title mark.\\
\texttt{stream\_overlay} & Facecam, donation, event, creator status, or live-broadcast overlay.\\
\texttt{other\_hud} & Visible UI with no fitting concrete class.\\
\bottomrule
\end{tabular}
\end{table*}

\section{Game2World Dataset Construction}
\label{app:dataset-construction}

This section describes the complete construction pipeline for the Game2World dataset. It covers the curation and standardization of clean gameplay clips, the selection and spatial organization of verified UI assets, and the synthesis of persistent and transient UI elements. Together, these stages produce diverse and temporally consistent paired data for gameplay UI removal and downstream world-model training.

\subsection{Clean Gameplay Corpus Construction}
\label{app:clean-corpus}

This subsection explains how raw gameplay footage is segmented, represented, diversity-sampled, temporally deduplicated, and standardized. The resulting corpus provides visually diverse and uniformly formatted clean gameplay clips for subsequent UI synthesis and model training.

\paragraph{Temporal Standardization and Visual Representation}

To provide uniform training examples, source gameplay is divided into complete, non-overlapping five-second windows. The central frame of each window is used as a compact visual representative for diversity analysis, balancing scene coverage against the cost of processing every frame. Before feature extraction, this representative frame is resized and center-cropped to $224\times224$, making comparisons less sensitive to the source recording resolution and aspect ratio.

\paragraph{Diversity-Aware Sampling}

We extract normalized CLIP ViT-B/16 image embeddings and organize the candidates with spherical $k$-means. Let $e_i\in\mathbb{R}^d$ denote an $\ell_2$-normalized embedding. The number of clusters is defined as
\begin{equation}
K=\min\left(K_{\max},N,\max(2,\lfloor4\sqrt{N}\rfloor)\right),
\end{equation}
where $K_{\max}=2048$. Centroids are initialized from at most 500,000 candidates, updated for 15 iterations, and re-normalized after each update.

To prevent visually dominant games or scenes from overwhelming the corpus, we sample candidates in a cluster-balanced round-robin manner rather than in proportion to cluster size. This produces at most 100,000 diverse candidates before temporal deduplication.

\paragraph{Temporal Redundancy Removal}

Gameplay videos often contain consecutive windows with nearly identical environments and camera states. We therefore compare temporally adjacent candidates from the same source and remove the later candidate when their cosine similarity exceeds $0.94$. Candidate pairs are processed from highest to lowest similarity, and a candidate participates in at most one removal pair.

\paragraph{Standardized Clip Format}

The selected videos undergo a quality check based on brightness, blur, and motion criteria and are then standardized to five seconds, 720p, and 30 FPS. The source aspect ratio is preserved during resizing, with padding applied when necessary. The resulting uniform format ensures that UI synthesis and model training operate on consistent spatial and temporal dimensions.

\subsection{UI Asset Selection and Persistent-UI Layout}
\label{app:asset-layout}

This subsection describes how verified UI assets are selected and arranged as persistent UI elements. It introduces the eligibility criteria, category-specific spatial priors, common layout formations, seeded position variation, collision handling, and motion models used to generate realistic yet diverse interface layouts.

\paragraph{Asset Eligibility and Aspect-Ratio Fallback}

We retain only assets that pass human verification and are marked suitable for rendering. Assets are filtered by taxonomy, visual form, and role-specific aspect-ratio constraints. To avoid unnecessarily discarding small but unusual interface components, an asset that violates only the aspect-ratio constraint remains eligible when its fitted area occupies at most 4\% of the frame.

Whenever the original screen position is available, its normalized center and scale serve as the initial placement prior. Otherwise, the renderer uses category-specific spatial anchors. This preserves realistic layout statistics while allowing controlled variation across synthesized clips.

\paragraph{Persistent UI Role Priors}

Table~\ref{tab:persistent-role-specs} summarizes the spatial and motion priors used for persistent UI roles. Maximum dimensions and motion amplitudes are expressed as fractions of the output frame.

\begin{table*}[t]
    \centering
    \scriptsize
    \begin{tabular*}{\textwidth}{
        @{\extracolsep{\fill}}
        llcccccc
        @{}
    }
        \toprule
        \textbf{Role}
        & \textbf{Taxonomy}
        & \textbf{Aspect}
        & $w_{\max}$
        & $h_{\max}$
        & \textbf{Priority}
        & $p_{\mathrm{motion}}$
        & \textbf{Motion frac. $(x,y)$} \\
        \midrule
        \texttt{minimap}
        & \texttt{map\_radar}
        & 0.65--1.65
        & 0.185
        & 0.290
        & 80
        & 0.25
        & (0.0012, 0.0008) \\
        \texttt{compass\_bar}
        & \texttt{navigation\_compass}
        & 1.80--18.00
        & 0.390
        & 0.065
        & 70
        & 0.42
        & (0.0018, 0.0008) \\
        \texttt{health}
        & \texttt{player\_status}
        & 1.35--7.50
        & 0.245
        & 0.110
        & 85
        & 0.20
        & (0.0010, 0.0007) \\
        \texttt{stamina}
        & \texttt{player\_status}
        & 1.35--8.50
        & 0.205
        & 0.060
        & 84
        & 0.24
        & (0.0010, 0.0008) \\
        \texttt{weapon\_bar}
        & \texttt{inventory}
        & 1.45--8.50
        & 0.360
        & 0.175
        & 75
        & 0.28
        & (0.0012, 0.0007) \\
        \texttt{crosshair}
        & \texttt{crosshair}
        & 0.40--2.50
        & 0.055
        & 0.085
        & 100
        & 0.55
        & (0.0018, 0.0018) \\
        \texttt{objective\_marker}
        & \texttt{world\_context}
        & 0.25--4.00
        & 0.100
        & 0.120
        & 55
        & 0.92
        & (0.0120, 0.0070) \\
        \texttt{ammo\_counter}
        & \texttt{weapon\_ammo}
        & 0.40--8.00
        & 0.130
        & 0.090
        & 65
        & 0.32
        & (0.0014, 0.0010) \\
        \bottomrule
    \end{tabular*}
    \caption{Spatial and motion priors for persistent UI roles.}
    \label{tab:persistent-role-specs}
\end{table*}

The default anchors are top-left for the minimap, top-center for the compass, bottom-left for health and stamina, bottom-center for the weapon bar, center for the crosshair, a free screen-space position for the objective marker, and bottom-right for ammunition. Recorded positions override these anchors when available.

\paragraph{Formation Presets}

We use three layout families that reflect common gameplay-interface organizations: an adventure layout with a top-left minimap and bottom-left status panel, a tactical layout with a top-right minimap and top-left status panel, and an RPG layout with a bottom-left minimap and top-right status panel. The weapon bar remains near the bottom center in all three families. These layouts provide fallback anchors and category-specific size priors; recorded source coordinates remain preferred whenever available.

\paragraph{Seeded Position Jitter and Collision Resolution}

The 720p five-second profile applies independent seeded jitter to every selected persistent widget:
\begin{align}
\Delta x &\sim \mathcal{U}(-0.018W,0.018W), \notag\\
\Delta y &\sim \mathcal{U}(-0.018H,0.018H).
\end{align}
At 1280$\times$720, these ranges correspond to maximum continuous offsets of 23.04 horizontal pixels and 12.96 vertical pixels before integer placement.

Persistent widgets are processed in decreasing layout priority. Visible alpha support is defined by an alpha threshold of 8. Collision rectangles include both a base padding and the role's complete motion envelope. For two rectangles $A$ and $B$, overlap is measured as
\begin{equation}
\rho(A,B)=\frac{|A\cap B|}{\min(|A|,|B|)}.
\end{equation}
The default acceptable threshold is $0.14$. Candidate displacements include small frame-relative offsets and object-size-relative offsets. The chosen position minimizes the squared-overlap cost plus the normalized displacement from the preferred position. A lower-priority widget is suppressed only when no candidate satisfies the overlap threshold.

For a moving static widget, horizontal and vertical amplitudes are drawn from the role-specific fractions, phases are sampled uniformly from $[0,2\pi)$, and periods are sampled from 5--9 seconds horizontally and 6--11 seconds vertically. The selected placement, jitter, collision score, suppression state, and motion parameters are retained in the synthesis record.

\subsection{Synthetic UI Composition}
\label{app:composition}

This subsection presents the complete UI composition process. It describes the global style presets, persistent-widget sampling policy, transient-event distribution, asset-taxonomy mixtures, event geometry and timing, and animation models used to synthesize diverse and temporally coherent gameplay interfaces.

\paragraph{Style Presets}

With \texttt{HUD\_STYLE=auto}, the renderer selects one of five presets through a seed-specific uniform choice. Each preset changes the formation, global UI scale, opacity scale, selected role scales, and transient-event density.

\begin{table*}[t]
    \centering
    \small
    \begin{tabular}{
        @{}
        llcccp{0.32\textwidth}
        @{}
    }
        \toprule
        \textbf{Style}
        & \textbf{Formation}
        & \textbf{Opacity scale}
        & \textbf{UI scale}
        & \textbf{Transient density}
        & \textbf{Role-specific scales} \\
        \midrule
        Adventure
        & Adventure
        & 0.94
        & 1.00
        & 0.95
        & minimap 1.06, health 1.06, weapon bar 1.00 \\
        Tactical
        & Tactical
        & 1.00
        & 0.96
        & 1.20
        & minimap 1.02, health 0.92, weapon bar 1.03, crosshair 1.14 \\
        RPG
        & RPG
        & 0.92
        & 1.04
        & 1.10
        & minimap 1.08, health 1.08, weapon bar 1.00 \\
        Survival
        & Adventure
        & 0.88
        & 0.92
        & 0.75
        & minimap 0.95, health 0.98, weapon bar 0.90, crosshair 0.92 \\
        Arena
        & Tactical
        & 1.02
        & 1.02
        & 1.35
        & minimap 0.98, health 0.86, weapon bar 1.08, crosshair 1.22 \\
        \bottomrule
    \end{tabular}
    \caption{UI style presets in the released renderer.}
    \label{tab:hud-styles}
\end{table*}

The global effective opacity is
\begin{equation}
\alpha_{\mathrm{HUD}}
=
\mathrm{clip}(0.94\,s_{\alpha},0,1),
\end{equation}
where $s_{\alpha}$ is the style opacity multiplier. Role size limits are multiplied by the global style scale and the role-specific scale and clipped to at most 0.92 of the corresponding frame dimension.

\paragraph{Persistent Widget Policy}

Each persistent taxonomy is sampled with an independent seed stream. The 720p five-second profile uses the probabilities in Table~\ref{tab:static-spawn}. Since every persistent category has a maximum count of one, the expected number is 4.00 widgets per video. There is no hard per-video quota, so sparse and dense realizations remain possible.

\begin{table}[t]
    \centering
    \small
    \begin{tabular}{lcc}
        \toprule
        \textbf{Taxonomy}
        & \textbf{Probability}
        & \textbf{Maximum} \\
        \midrule
        \texttt{map\_radar}           & 0.68 & 1 \\
        \texttt{player\_status}       & 0.72 & 1 \\
        \texttt{inventory}            & 0.58 & 1 \\
        \texttt{navigation\_compass}  & 0.48 & 1 \\
        \texttt{crosshair}            & 0.64 & 1 \\
        \texttt{weapon\_ammo}         & 0.54 & 1 \\
        \texttt{world\_context}       & 0.36 & 1 \\
        \bottomrule
    \end{tabular}
    \caption{Persistent element spawn policy.}
    \label{tab:static-spawn}
\end{table}

\paragraph{Total Popup Count and Event Categories}

For five-second clips, the total number $M$ of popup events is sampled before choosing event categories:
\begin{equation}
P(M=0,1,2,3,4,5)
=
(0.14,0.43,0.28,0.10,0.04,0.01).
\end{equation}
The expected total is exactly 1.5 events. Conditional on $M$, categories are drawn without exceeding their configured maximum counts, with weights proportional to the probabilities in Table~\ref{tab:popup-spawn}. The more general default profile without an explicit total-count distribution instead performs independent Bernoulli draws and has an expected count of 5.50 events per video.

\begin{table}[t]
    \centering
    \scriptsize
    \setlength{\tabcolsep}{3pt}
    \begin{tabular}{
        @{}
        >{\raggedright\arraybackslash}p{0.43\columnwidth}
        cc
        @{}
    }
        \toprule
        \textbf{Event category}
        & \textbf{Weight}
        & \textbf{Max.} \\
        \midrule
        \texttt{world\_context}       & 0.65 & 3 \\
        \texttt{objective\_status}    & 0.45 & 2 \\
        \texttt{interaction\_prompt}  & 0.55 & 2 \\
        \texttt{screen\_notice}       & 0.60 & 3 \\
        \texttt{ability\_status}      & 0.35 & 2 \\
        \texttt{teammate\_roster}     & 0.20 & 2 \\
        \texttt{match\_status}        & 0.45 & 2 \\
        \bottomrule
    \end{tabular}
    \caption{Popup event-category policy. Under the five-second profile, these values serve as relative weights after the total popup count is drawn.}
    \label{tab:popup-spawn}
\end{table}

Independent deterministic random streams are used for composition, asset selection, timing, and animation. This decoupling preserves reproducibility and prevents a change in one sampling component from unintentionally altering the others.

\paragraph{Popup Asset-Taxonomy Mixture}

The event category determines placement, timing, size, and animation behavior, while the visual asset may be drawn from a related taxonomy. Table~\ref{tab:taxonomy-mix} gives the default mixture. The synthesis record retains both the logical event category and the selected asset taxonomy so that the generated supervision remains interpretable.

\begin{table*}[t]
    \centering
    \small
    \begin{tabular}{
        @{}
        l
        p{0.73\linewidth}
        @{}
    }
        \toprule
        \textbf{Event category}
        & \textbf{Asset taxonomy mixture} \\
        \midrule
        \texttt{objective\_status}
        & \texttt{objective\_status}: 0.75;
          \texttt{subtitle}: 0.25 \\
        \texttt{interaction\_prompt}
        & \texttt{interaction\_prompt}: 0.75;
          \texttt{menu\_controls}: 0.25 \\
        \texttt{screen\_notice}
        & \texttt{screen\_notice}: 0.65;
          \texttt{performance\_debug}: 0.15;
          \texttt{watermark}: 0.10;
          \texttt{other\_hud}: 0.10 \\
        \texttt{ability\_status}
        & \texttt{ability\_status}: 0.75;
          \texttt{vehicle\_status}: 0.25 \\
        \texttt{teammate\_roster}
        & \texttt{teammate\_roster}: 0.60;
          \texttt{chat\_comms}: 0.25;
          \texttt{stream\_overlay}: 0.15 \\
        \texttt{match\_status}
        & \texttt{match\_status}: 1.00 \\
        \bottomrule
    \end{tabular}
    \caption{Default mapping from event categories to source-asset taxonomies.}
    \label{tab:taxonomy-mix}
\end{table*}

\paragraph{Transient Event Geometry and Timing}

Table~\ref{tab:transient-specs} lists the released event profiles. The anchor is selected from the profile's candidate set. Event durations are sampled from the listed intervals and clipped to the available clip duration. The maximum size is expressed as a fraction of the video frame.

\begin{table*}[t]
    \centering
    \scriptsize
    \begin{tabular}{
        lccccc
        p{0.30\linewidth}
    }
        \toprule
        \textbf{Category}
        & $w_{\max}$
        & $h_{\max}$
        & \textbf{Duration (s)}
        & \textbf{Fade (s)}
        & \textbf{Opacity}
        & \textbf{Anchors} \\
        \midrule
        \texttt{objective\_status}
        & 0.340
        & 0.105
        & 2.3--5.8
        & 0.35
        & 0.88
        & lower-center, upper-center, center-right \\
        \texttt{interaction\_prompt}
        & 0.185
        & 0.120
        & 1.4--3.8
        & 0.22
        & 0.86
        & center-right, center-left, near-center, upper-center \\
        \texttt{screen\_notice}
        & 0.300
        & 0.130
        & 2.0--4.5
        & 0.28
        & 0.91
        & top-right stack, top-left stack \\
        \texttt{ability\_status}
        & 0.145
        & 0.100
        & 4.0--9.5
        & 0.55
        & 0.90
        & lower-left stack, upper-right stack, above weapon \\
        \texttt{teammate\_roster}
        & 0.230
        & 0.150
        & 3.2--7.0
        & 0.40
        & 0.88
        & left-mid, right-mid, top-left stack \\
        \texttt{match\_status}
        & 0.155
        & 0.090
        & 1.0--2.6
        & 0.20
        & 0.88
        & near-center, center-right, upper-center \\
        \bottomrule
    \end{tabular}
    \caption{Transient category profiles. The implementation also stores category-specific aspect-ratio bounds and drift fractions.}
    \label{tab:transient-specs}
\end{table*}

The corresponding drift fractions are $(0.004,0.003)$, $(0.010,0.006)$, $(0.002,0.004)$, $(0.002,0.002)$, $(0.004,0.003)$, and $(0.006,0.014)$, following the category order in Table~\ref{tab:transient-specs}.

\paragraph{Popup Animation Distribution}

Each event independently draws one of six animation archetypes:

\begin{table*}[t]
    \centering
    \small
    \begin{tabular}{
        l
        c
        p{0.67\linewidth}
    }
        \toprule
        \textbf{Animation}
        & \textbf{Weight}
        & \textbf{Implementation} \\
        \midrule
        \texttt{fade}
        & 0.22
        & Smooth, independently sampled entrance and exit fades. \\
        \texttt{instant}
        & 0.13
        & Immediate appearance and hard disappearance. \\
        \texttt{slide}
        & 0.16
        & Horizontal entrance with a smaller exit displacement and opacity ramp. \\
        \texttt{wipe\_trail}
        & 0.18
        & Irregular left-to-right scanline reveal/removal with deterministic high-frequency modulation. \\
        \texttt{pop}
        & 0.09
        & Ease-out-back scale overshoot, settle, and shrink. \\
        \texttt{border\_arc}
        & 0.22
        & Enter from one frame border, traverse a noisy parabolic arc, and exit through a different border. \\
        \bottomrule
    \end{tabular}
    \caption{Popup animation weights and behaviors.}
    \label{tab:popup-animation}
\end{table*}

Entrance and exit durations are capped at 32\% of the event duration. Before applying this cap, the sampled entrance/exit ranges are 0.18--0.58/0.18--0.62 seconds for \texttt{fade}, 0.26--0.72/0.16--0.48 seconds for \texttt{slide}, 0.34--0.82/0.28--0.70 seconds for \texttt{wipe\_trail}, and 0.18--0.46/0.12--0.38 seconds for \texttt{pop}. The \texttt{instant} and \texttt{border\_arc} animations have zero entrance and exit fade durations.

For \texttt{border\_arc}, entry and exit borders are sampled independently subject to being different. Arc height is sampled from 0.14--0.34 of the shorter frame side, normal-direction noise amplitude from 0.010--0.032, and noise frequency from 1.7--3.8 cycles. All parameters are recorded in the metadata.

\section{Additional Dataset Statistics and Details}
\label{app:dataset-details}
Each benchmark case is defined by a consistent case specification that links the UI-containing source, an optional clean reference, temporal annotations, and the video properties needed for timestamp-aligned evaluation.

\begin{table}[t]
\centering
\caption{Information required to define a benchmark case.}
\label{tab:manifest-fields}
\small
\setlength{\tabcolsep}{4pt}
\renewcommand{\arraystretch}{1.08}
\begin{tabular}{@{}>{\raggedright\arraybackslash}p{0.31\columnwidth}>{\raggedright\arraybackslash}p{0.61\columnwidth}@{}}
\toprule
\textbf{Field} & \textbf{Purpose}\\
\midrule
Case identifier & Uniquely associates a model output with its source example.\\
Dataset split & Distinguishes synthetic and in-the-wild evaluation cases.\\
Source video & UI-containing video supplied to the removal model.\\
Clean reference & Available only for synthetic cases and retained for controlled analysis.\\
Annotation & Frame-level synthetic boxes or human-annotated in-the-wild trajectories.\\
Video metadata & Width, height, frame rate, frame count, and annotation format.\\
\bottomrule
\end{tabular}
\end{table}

For synthetic cases, the evaluated source is always the UI-overlaid video. The paired clean video is retained for controlled analysis but is not shown to the primary evaluator. In-the-wild cases have no clean reference. A stable case identifier ensures that each edited result is compared with the intended source and annotation.

\section{Video-Conditioned Aesthetic Matching}
\label{app:aesthetic}

Aesthetic matching is optional and enabled in the default synthesis profile. Without it, assets are sampled uniformly from the taxonomy-compatible pool. When enabled, the renderer constructs a visual appearance profile for the clean video and uses weighted sampling to favor compatible assets while retaining diversity rather than always selecting the single highest-scoring candidate.

\subsection{Diverse keyframe selection}

For a requested $K$ keyframes, the renderer uniformly probes
\begin{equation}
P=\min(N_{\mathrm{frames}},\max(2K,10))
\end{equation}
video frames. It computes an appearance embedding for every probe. The first retained frame is the one with maximum cosine similarity to the normalized median embedding. Remaining frames are selected by farthest-point sampling, each time maximizing distance to the nearest selected embedding. The default retained keyframe count is eight.

\subsection{Appearance descriptor}

Transparent asset pixels are ignored using alpha weights. The hand-crafted descriptor concatenates normalized histograms of HSV hue, saturation, and value; Lab $a$ and $b$; gray-level intensity; gradient orientation; and log gradient magnitude. It further includes normalized means and scalar statistics for value, saturation, edge magnitude, edge density, and the 10th/90th value percentiles. Every group and the final concatenation are $\ell_2$-normalized.

An optional EfficientNet-B0 representation complements the hand-crafted descriptor when available. Its global-average-pooled feature is normalized before similarity computation; otherwise, the hand-crafted appearance descriptor is used alone.

\subsection{Similarity and sampling}

For an asset descriptor $a$ and a set of keyframe descriptors $V$, profile similarity is
\begin{equation}
s(a,V)=0.70\,\mathrm{median}(Va)+0.30\,\max(Va).
\end{equation}
When both visual and EfficientNet features are available, the final score is
\begin{equation}
s_{\mathrm{final}}=0.68s_{\mathrm{visual}}+0.32\frac{s_{\mathrm{neural}}+1}{2}.
\end{equation}
Given candidate scores with range $r$, weighted sampling uses a softmax-like temperature
\begin{equation}
\tau=\max(0.025,0.24r).
\end{equation}
For each role, the renderer retains the selection score and the highest-ranked alternatives. This record supports reproducibility and makes it possible to inspect whether appearance matching selected assets consistent with the target gameplay.

\section{Structured Status and Inventory Rendering}
\label{app:structured-ui}

\subsection{Player-status panels}

When a status asset includes internal bar geometry, the renderer animates the filled portion over time. Assets without this internal structure are composited directly, and a procedural status panel is used only when no suitable asset is available. This preserves real interface appearance whenever possible while still supporting temporally varying status signals.

For dynamic assets, health and stamina values evolve smoothly as
\begin{align}
h(t)&=\operatorname{Smooth}(t;9.0,0.2,0.42,0.96),\\
s(t)&=\operatorname{Smooth}(t;5.0,1.6,0.25,0.92),
\end{align}
where \texttt{Smooth} combines a primary sinusoid and a lower-frequency secondary sinusoid before clipping to $[0,1]$. Filled regions preserve the original panel color; unfilled saturated regions are darkened to 30\% intensity. If no suitable asset exists, the renderer draws procedural \texttt{HP} and \texttt{STM} bars.

\subsection{Inventory and hotbar components}

Separated inventory assets are stored as a background bar plus reusable item icons. The separator requires a high-confidence empty slot from the same repeated slot row, tiles that local background to reconstruct a clean bar, and extracts conservative icon differences. Ambiguous rows are recorded as skipped instead of entering the active pool.

At render time, the background bar is selected first. Candidate icons must be at least $8\times8$ pixels and have aspect ratio between 0.15 and 5.8. At most 120 visually valid icons are retained for a video. A new deterministic inventory state is sampled every 3.2--7.8 seconds. Slot occupancy is uniform in $[0.62,1.0]$; each icon is fitted to 78\% of the slot dimensions, receives small seeded position jitter, and is composited at opacity 0.96. The frame-level annotation exporter emits these icons as child instances of the composed weapon bar.

\section{Synthetic Supervision and Reproducibility Records}
\label{app:renderer-output}

Each synthesized example contains an aligned clean video, UI-overlaid video, binary UI mask, and structured metadata. The clean video serves as the reconstruction target and the overlaid video as the model input. Optional visual summaries support quality control, and appearance comparisons are retained when aesthetic matching is enabled.

The synthesis record contains the taxonomy version, random seed, selected layout and style, realized persistent and transient elements, collision-adjusted positions, opacity, video dimensions and duration, asset provenance, status and inventory rendering modes, event timing, animation parameters, and optional appearance-matching diagnostics. Its purpose is to make every generated example reproducible and to support deterministic reconstruction of frame-level annotations.

The reported 720p five-second profile uses UI opacity 0.94, automatic style and layout selection, dynamic status and skill components, transient events, a collision threshold of 0.14, position jitter $(0.018,0.018)$, and an aspect-ratio fallback area of 0.04. Aesthetic matching is disabled in this profile.

\section{Frame-Level Synthetic Ground Truth}
\label{app:frame-bbox}

Frame-level annotations are reconstructed deterministically from the synthesis metadata. All boxes use clipped, half-open pixel coordinates $[x_1,y_1,x_2,y_2]$, so their width and height are $x_2-x_1$ and $y_2-y_1$.

Each annotation record contains the video dimensions, frame rate, frame count, coordinate convention, alpha-visibility threshold, an instance table, and a per-frame list of visible objects. The instance table stores identity, semantic category, role, asset provenance, and parent-child relations once per video. Each frame then references the visible instances and their current boxes.

Persistent boxes incorporate layout jitter and any sampled motion trajectory. Transient boxes reproduce entrance, exit, drift, trail, and border-arc animation, including off-screen clipping. Bounding boxes are tightened to alpha support using threshold 8. Partially visible transient elements additionally record their visible fraction. Separated inventory icons are represented as child instances of the composed inventory or weapon panel. Procedurally rendered text is included in the pixel mask but is not treated as a separate asset instance.

This representation provides exact frame-aligned supervision while preserving temporal identity across appearances of the same UI component.

\section{Additional \method Implementation Details}
\label{app:model-details}

\method{} follows the Kiwi-Edit architecture used in the main paper. A pretrained MLLM encodes sampled source frames and the removal instruction. Learnable latent queries extract task-relevant visual-semantic information through cross-attention, and a connector projects these features into the video DiT conditioning space:
\begin{equation}
c=P\!\left(\operatorname{CrossAttn}\left(Q,E_{\mathrm{MLLM}}(x_{\mathrm{src}},y)\right)\right).
\end{equation}
The source-video latent is injected into the noisy target stream using a timestep-dependent residual:
\begin{equation}
h_t=\operatorname{PE}(z_t)+\gamma(t)\operatorname{PE}_{\mathrm{src}}\!\left(\operatorname{VAE}(x_{\mathrm{src}})\right).
\end{equation}
Training uses the standard flow-matching objective
\begin{equation}
\mathcal{L}_{\mathrm{flow}}=
\mathbb{E}_{t,z_0,z_1,c}
\left[\left\|v_\theta(z_t,t,c)-(z_1-z_0)\right\|_2^2\right],
\end{equation}
where $z_1$ is the clean target latent and $z_0$ is Gaussian noise.

The model is initialized from the Stage-2 instruction-editing checkpoint of Kiwi-Edit. The MLLM and connector are frozen. LoRA with rank 64 is applied to the DiT. Training uses eight H200 GPUs, a global batch size of 32, learning rate $10^{-4}$, 1,400 optimization steps, and 720p paired videos.

\section{Prompts Used in the Reported Evaluation}
\label{app:prompts}

\subsection{Video-edit generation prompts}

Generation prompts are consumed by the video-editing model during inference and are not visible to the evaluator. The frozen timeline-correct protocol records the following exact strings for the two Kiwi runs:

\begin{table}[h]
\centering
\small
\begin{tabular}{@{}p{0.25\columnwidth}p{0.67\columnwidth}@{}}
\toprule
\textbf{Run} & \textbf{Exact generation prompt}\\
\midrule
General Video Edit Models & \texttt{This is a gameplay video. Please remove the UI and UI elements from it without affecting the gameplay footage.}\\
\method & \texttt{Remove the gameplay UI from the video.}\\
\bottomrule
\end{tabular}
\caption{Exact video-editing prompts recorded in the frozen evaluation protocol.}
\label{tab:generation-prompts}
\end{table}

The frozen protocol specifies that each model inference report records the generation prompt, inference steps, seed, output FPS, and output resolution. These fields are not provided to the judge.

\subsection{Complete MLLM evaluation prompt}
\label{app:judge-prompt}

The evaluator dynamically substitutes only \texttt{\{IMAGE\_COUNT\}} and \texttt{\{EVIDENCE\_MAPPING\_JSON\}}. The frozen Terra judge prompt is reproduced verbatim below.

\begin{lstlisting}[style=g2wcode]
You are a strict visual evaluator for gameplay HUD/UI removal.

You receive {IMAGE_COUNT} evidence images from one video in the exact order
listed below.

There are two attached image kinds:

1. "overall": one canonical-resolution ORIGINAL/EDITED full-frame pair for a
   timestamp. If layout is vertical, ORIGINAL is top and EDITED is bottom. If
   horizontal, ORIGINAL is left and EDITED is right. Red rectangles only
   outline annotated regions and were added after editing.
2. "element": one detail sheet for exactly one
   (sample_id, element_id). Every sheet has a labeled canonical-resolution
   CONTEXT pair expanded beyond the bbox. Small targets also have a labeled
   EXACT TARGET pair that repeats only the exact bbox. Each panel independently
   contains ORIGINAL and EDITED sides in its mapped pair layout; never compare
   one panel against the other. The exact panel is a locator, not an edited copy
   of the context panel. Both sides have the same pixel scale.

All overall images are attached first in timestamp order, followed by all
element images in sample/element order. Inspect every overall image explicitly
for frame preservation before judging the element crops.

Before any bbox mapping or crop extraction, both ORIGINAL and EDITED
videos are independently scaled to the same canonical canvas recorded in the
evidence metadata. Corresponding crops therefore use the same pixel scale.
No crop is resized again after extraction. Neutral black padding/separators and
E-label headers are renderer canvas, not video content.
No box, label, or text is drawn over detail-pair video pixels.

For every listed element, decide:
- "removed": the specific original HUD/UI element is no longer visible in the
  edited crop;
- "present": the element or a recognizable residual fragment is still visible;
- "uncertain": the evidence is genuinely insufficient.

Track the identity and pixels of the SPECIFIC original-side element, not merely
its semantic category. Before choosing "removed", identify at least one
distinctive anchor in the original (text strokes, icon contour, bar ornament,
reticle arms, or another stable component) and verify that anchor is absent
from the edited side. Respect the stated top/bottom or left/right orientation;
never compare two original regions. A multi-part UI element is "present" if
any recognizable component survives, even when its other components are gone.
Faint, blurred, partially readable, or fragmented remnants of the same
text/icon/shape also count as "present".

Annotated elements can overlap pre-existing game UI or other annotated
elements. The edited crop may therefore still contain unrelated text, bars,
icons, or native HUD. That alone is not evidence of "present". Use "present"
only when the same distinctive original graphic/text/shape, or a recognizable
fragment of it, survives. If that original-only layer is gone while unrelated
content remains, use "removed".

For a removed element, rate the edited fill:
- "none": plausible scene content with no visible removal artifact;
- "minor": small blur/seam/texture defect;
- "major": obvious hole, smear, copied patch, or destructive corruption;
- "uncertain": fill quality cannot be determined.
For present/uncertain elements use "not_applicable".
Ordinary underlying scene or native-game content is not an artifact, even when
it has a similar color or brightness to the removed overlay. Natural motion
blur, depth-of-field blur, texture softness, and codec softness are also not
artifacts by themselves. Mark blur/softness as an artifact only when it is a
localized editing defect inconsistent with adjacent edited content or has a
visible boundary/seam.

For every frame, rate preservation OUTSIDE all annotated UI regions:
"preserved", "minor_change", "major_change", or "invalid". Natural codec
differences are preserved. Large scene, geometry, camera, color, or content
changes are not preserved. At a matched timestamp, a substantially different
player/enemy position, pose, animation state, camera viewpoint, or scene timing
is a major change even when the room, level, or native UI looks similar. Do
not infer "preserved" merely because both panels depict the same game/location.
"invalid" means the comparison itself is unusable. Frame notes must cite a
concrete matched feature or concrete difference visible in both panels; avoid
generic notes such as "scene matches".

Judge each element independently. Do not infer success from the editing method,
the filename, or other elements. Return exactly one element record for every
listed (sample_id, element_id) and exactly one frame record per sample_id.
Confidence is for auditing only and must be between 0 and 1. Keep each note
short and visual.

Evidence mapping:
{EVIDENCE_MAPPING_JSON}
\end{lstlisting}

\section{HUD-Removal Evaluation Protocol}
\label{app:evaluation}

\subsection{Timestamp-based frame alignment}

Evaluation samples the source timeline at $f_{\mathrm{eval}}=2$ FPS:
\begin{equation}
t_k=\frac{k}{f_{\mathrm{eval}}}.
\end{equation}
The source index is
\begin{equation}
i_s(k)=\operatorname{round}(t_k f_s),
\end{equation}
and the edited index is independently remapped by timestamp:
\begin{equation}
i_e(k)=\min\left(N_e-1,\operatorname{round}(t_k f_e)\right),
\end{equation}
where $f_s$ and $f_e$ are the source and edited-video frame rates. Thus, a 30-FPS, 150-frame source is sampled at indices $0,15,\ldots,135$, while an 81-frame, 16-FPS edited result is sampled at $0,8,\ldots,72$. This avoids comparing identical integer frame indices at different physical times. A case fails when the edited duration ends more than 0.05 seconds before the final evaluation timestamp.

Before sampling, the evaluator validates source and edited-video dimensions, frame rates, frame counts, and durations. This prevents invalid comparisons caused by incomplete outputs or incompatible geometry; the allowed absolute aspect-ratio difference is at most 0.01.

\subsection{Canonical resize and box mapping}

Source and edited frames are independently scaled to the same canonical canvas using Lanczos interpolation before any box mapping or cropping. This is direct scaling rather than letterboxing. The default canonical canvas equals the annotation canvas, 1280$\times$720. Experiments at 1920$\times$1080 use that resolution consistently for every evaluated method, and results from different canonical canvases are not mixed in one formal comparison table.

For annotation canvas $W_a\times H_a$ and canonical canvas $W_c\times H_c$, a box $(x_1,y_1,x_2,y_2)$ is mapped as
\begin{align}
x'_1&=\left\lfloor x_1W_c/W_a\right\rfloor,
&y'_1&=\left\lfloor y_1H_c/H_a\right\rfloor,\\
x'_2&=\left\lceil x_2W_c/W_a\right\rceil,
&y'_2&=\left\lceil y_2H_c/H_a\right\rceil.
\end{align}
The mapped box is clipped to the canonical frame and guaranteed to have at least one pixel in each dimension. Source and edited frames therefore share exactly the same box coordinates and pixel scale.

For a mapped box of width $w$ and height $h$, context padding is
\begin{align}
p_x&=\max\left(8,\left\lceil w(1.5-1)/2\right\rceil\right),\\
p_y&=\max\left(8,\left\lceil h(1.5-1)/2\right\rceil\right).
\end{align}
The context crop is $[x_1-p_x,y_1-p_y,x_2+p_x,y_2+p_y]$ after clipping. If the longest side of the original annotation-space box is at most 96 pixels, the detail sheet also includes an exact-target pair. Using the annotation-space threshold prevents the panel set from changing when canonical resolution changes.

\subsection{Evidence attachments}

Each sampled timestamp produces one full-frame overview image. Landscape videos use a vertical pair with ORIGINAL above EDITED; portrait videos use a horizontal pair with ORIGINAL left of EDITED. Red boxes are drawn only on the overview. A black separator divides the two sides.

Each $(\texttt{sample\_id},\texttt{element\_id})$ produces one element detail sheet containing an identically scaled ORIGINAL/EDITED context pair. Small elements additionally receive an exact-target pair. Crops wider than tall are arranged vertically; crops taller than wide are arranged horizontally. Headers, labels, separators, and padding are placed outside video pixels.

All overview images are attached first in timestamp order, followed by all element sheets in sample/element order. The prompt includes a JSON evidence mapping containing attachment index, sample and element identifiers, pair orientation, dimensions, crop layout, and target-box coordinates relative to the crop.

\subsection{Judge output schema}

The judge must return exactly one element record for every expected pair and exactly one frame record per sampled timestamp:
\begin{lstlisting}[style=g2wcode]
{
  "elements": [
    {
      "sample_id": "f0000",
      "element_id": "E0",
      "removal": "removed",
      "artifact": "none",
      "confidence": 0.98,
      "note": "The original compass glyph is absent in the edited crop."
    }
  ],
  "frames": [
    {
      "sample_id": "f0000",
      "preservation": "preserved",
      "confidence": 0.97,
      "note": "Player pose, wall edges, and background geometry align."
    }
  ]
}
\end{lstlisting}
Allowed removal values are \texttt{removed}, \texttt{present}, and \texttt{uncertain}; artifact values are \texttt{none}, \texttt{minor}, \texttt{major}, \texttt{uncertain}, and \texttt{not\_applicable}; preservation values are \texttt{preserved}, \texttt{minor\_change}, \texttt{major\_change}, and \texttt{invalid}. Confidence must lie in $[0,1]$. Extra fields, missing keys, duplicate keys, and unexpected keys are rejected.

\subsection{Metric definitions}

For element $i$ in sampled frame $t$, strict UI removal is
\begin{equation}
r_{t,i}=\mathbb{1}[\texttt{removal}=\texttt{removed}],
\end{equation}
where \texttt{uncertain} is scored as failure. Clean removal is
\begin{equation}
c_{t,i}=\mathbb{1}[\texttt{removal}=\texttt{removed}\land\texttt{artifact}=\texttt{none}].
\end{equation}
Artifact-adjusted removal is zero unless removal succeeds; successful removals receive weight 1.0, 0.5, or 0.0 for \texttt{none}, \texttt{minor}, or \texttt{major}/\texttt{uncertain}, respectively.

For a frame with $N_t$ annotated elements,
\begin{align}
R_t&=\frac{1}{N_t}\sum_i r_{t,i}, &
C_t&=\frac{1}{N_t}\sum_i c_{t,i},\\
A_t&=\frac{1}{N_t}\sum_i a_{t,i}.&&
\end{align}
Frames without annotated elements do not enter removal metrics but remain in background preservation. For a video with evaluable frames $T_v$,
\begin{equation}
\operatorname{UIRemoval}_v=\frac{1}{|T_v|}\sum_{t\in T_v}R_t,
\end{equation}
with analogous clean and artifact-adjusted scores. Dataset aggregation is video-macro:
\begin{equation}
\operatorname{UIRemoval}_{\mathrm{dataset}}=\frac{1}{|V|}\sum_{v\in V}\operatorname{UIRemoval}_v.
\end{equation}
Thus the formal path is frame macro within each video, followed by video macro across the dataset.

Background preservation weights are 1.0 for \texttt{preserved}, 0.5 for \texttt{minor\_change}, and 0.0 for \texttt{major\_change} or \texttt{invalid}. They are averaged over sampled frames within a video and then macro-averaged over videos.

The implementation also reports box-micro removal, dataset-level frame-macro removal, uncertain rate, trajectory-complete removal, trajectory-complete clean removal, and evaluation coverage. The formal comparison uses UI removal success, clean UI removal success, artifact-adjusted removal, and background preservation, with synthetic and in-the-wild subsets reported separately.

\subsection{Judge execution and reproducibility}

For each video, the evaluator submits the complete evidence set and the prompt in a single read-only judge call. The call uses the frozen model and reasoning setting, validates the response against the exact JSON schema, and records both the raw response and the parsed judgment. Parallel evaluation changes only how many videos are processed concurrently; it does not change the one-call-per-video rule.

The protocol uses one frozen stochastic judgment per video. Resolution audits or evaluator-calibration studies must predeclare repetition or consensus rules and must not select the best result from repeated calls.

\includepdf[pages=-]{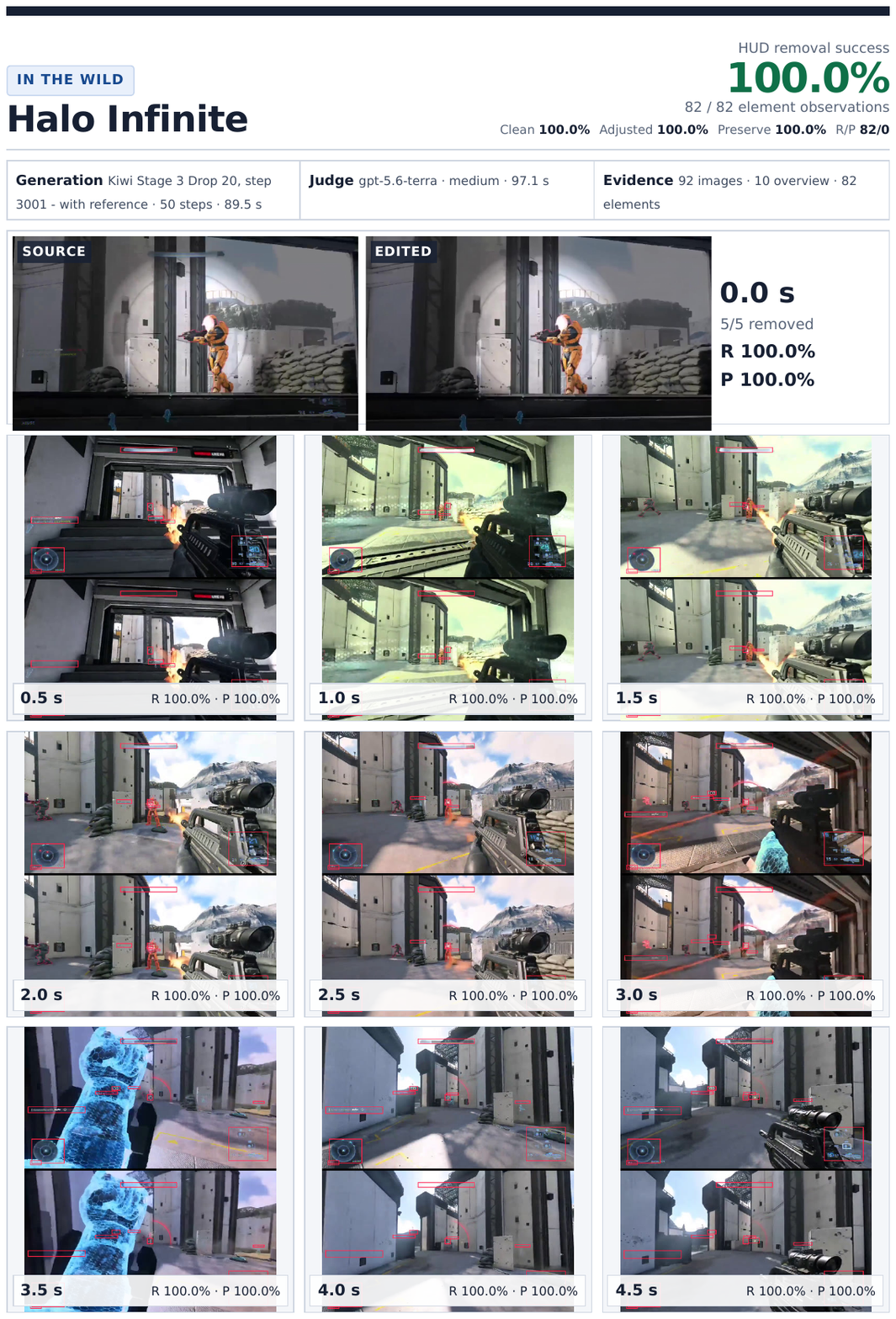}

\includepdf[pages=-]{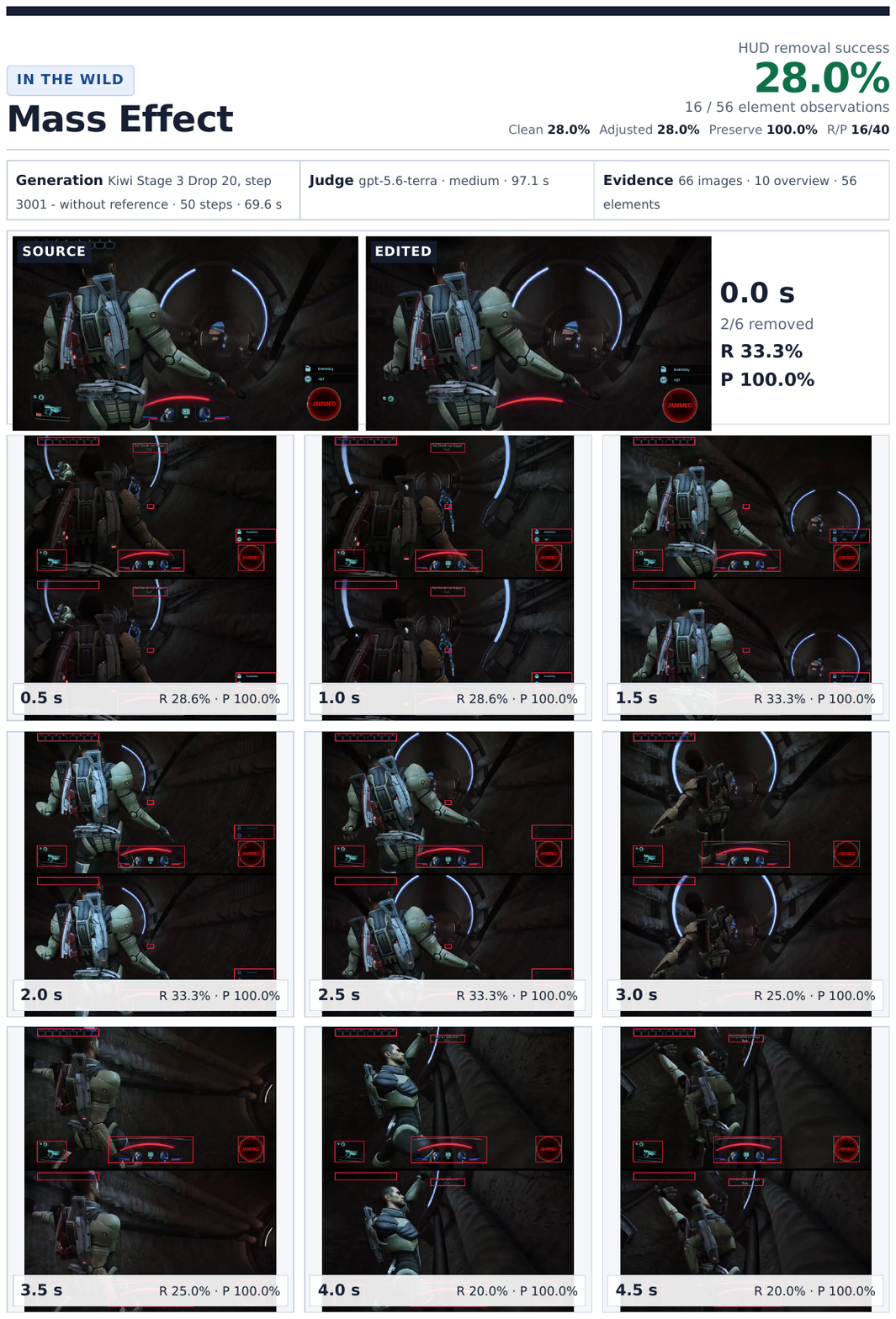}